\documentclass[final,5p,times,twocolumn]{elsarticle}

\usepackage{amsmath} 
\usepackage{amssymb} 
\usepackage{leftidx}
\usepackage{subfigure}
\usepackage{graphics}
\usepackage{comment}
\usepackage {color}
\usepackage{cuted}
\usepackage{multirow}
\usepackage{comment}
\usepackage[table,xcdraw]{xcolor}
\usepackage{titlesec}

\journal{Knowledge-Based Systems}

\begin{document}

\begin{frontmatter}



\title{DFMSD: Dual Feature Masking Stage-wise Knowledge Distillation for Object Detection}


\author[1]{Zhourui Zhang}
\ead{zzrnnupg@nnu.edu.cn}

\author[1]{Jun Li \corref{cor1}}
\ead{lijuncst@njnu.edu.cn}

\author[2]{Zhijian Wu}
\ead{zjwu_97@stu.ecnu.edu.cn}

\author[3]{Jifeng Shen}
\ead{shenjifeng@ujs.edu.cn}

\author[1]{Jianhua Xu}
\ead{xujianhua@njnu.edu.cn}

\address[1]{School of Computer and Electronic Information, Nanjing Normal University, Nanjing 210023, China}
\address[2]{School of Data Science and Engineering, East China Normal University, Shanghai 200062, China}
\address[3]{School of Electrical and Information Engineering, Jiangsu University, Zhenjiang 212013, China}

\cortext[cor1]{Corresponding author}


\begin{abstract}

In recent years, current mainstream feature masking distillation methods mainly function by reconstructing selectively masked regions of a student network from the feature maps of a teacher network. In these methods, attention mechanisms can help to identify spatially important regions and crucial object-aware channel clues, such that the reconstructed features are encoded with sufficient discriminative and representational power similar to teacher features. However, previous feature-masking distillation methods mainly address homogeneous knowledge distillation without fully taking into account the heterogeneous knowledge distillation scenario. In particular, the huge discrepancy between the teacher and the student frameworks within the heterogeneous distillation paradigm is detrimental to feature masking, leading to deteriorating reconstructed student features. In this study, a novel dual feature-masking heterogeneous distillation framework termed DFMSD is proposed for object detection. More specifically, a stage-wise adaptation learning module is incorporated into the dual feature-masking framework, and thus the student model can be progressively adapted to the teacher models for bridging the gap between heterogeneous networks. Furthermore, a masking enhancement strategy is combined with stage-wise learning such that object-aware masking regions are adaptively strengthened to improve feature-masking reconstruction. In addition, semantic alignment is performed at each Feature Pyramid Network (FPN) layer between the teacher and the student networks for generating consistent feature distributions. Our experiments for the object detection task demonstrate the promise of our approach, suggesting that DFMSD outperforms both the state-of-the-art heterogeneous and homogeneous distillation methods.

\end{abstract}




\end{frontmatter}


\section{Introduction}\label{sec1}


It is well-known that knowledge distillation (KD) can help transfer knowledge from a complex model (teacher) to a compact network (student), so that the latter can achieve improved performance at a much lower cost. It is considered to be an effective means of model compression for a variety of downstream tasks including object detection and semantic segmentation~\cite{r2,r5,r11,r64,r74,r75}. Primarily focusing on the output head of the network, early distillation algorithms aim at transferring implicit knowledge learned in the complex teacher network to the lightweight student model. This distillation scheme is also known as logit-based classification distillation~\cite{r2,r5,r75,r17}. In addition, the feature-based distillation approach has received increasing attention. It helps the student network to mimic feature maps from the teacher model in the distillation process, allowing the generated student features to enjoy improved representational capability~\cite{r4,r76}. More recently, a popular distillation paradigm has emerged as feature-masking distillation. In contrast to feature distillation in which the student's feature directly mimics the counterpart of the teacher~\cite{r77,r78}, feature-masking distillation operates by masking selective regions of the student feature map and reconstructing the masked regions for distillation~\cite{r19}. In this sense, feature-masking distillation essentially reconstructs the transferred knowledge from the teacher instead of transferring knowledge directly. Consequently, it can help the student learn better from the teacher. In particular, recent efforts are devoted to taking advantage of feature attention for uncovering object-aware spatially important regions and channel-wise clues such that the student features are reconstructed with sufficient descriptive power comparable to teacher features~\cite{r101}. As a result, this attention-directed feature masking strategy enormously contributes to improving the performance of the student model~\cite{r79}.

Although dramatic progress has been made in recent years, most feature-masking distillation methods are developed mainly to address homogeneous distillation, which assumes that teacher and student models share roughly similar structures except that the former usually adopts a stronger backbone. For example, RetinaNet-ResNet101~\cite{r8} and RetinaNet-ResNet50~\cite{r8} are used as the teacher and the student model, respectively, within the homogeneous distillation framework~\cite{r8,r29}. They fail to fully take into account heterogeneous distillation scenario which is more challenging due to significant diversity of the teacher and the student frameworks~\cite{r80}.  

\begin{figure}
\centering
\subfigure[Faster R-CNN]{
\includegraphics[width=1.4in,height=1.1in]{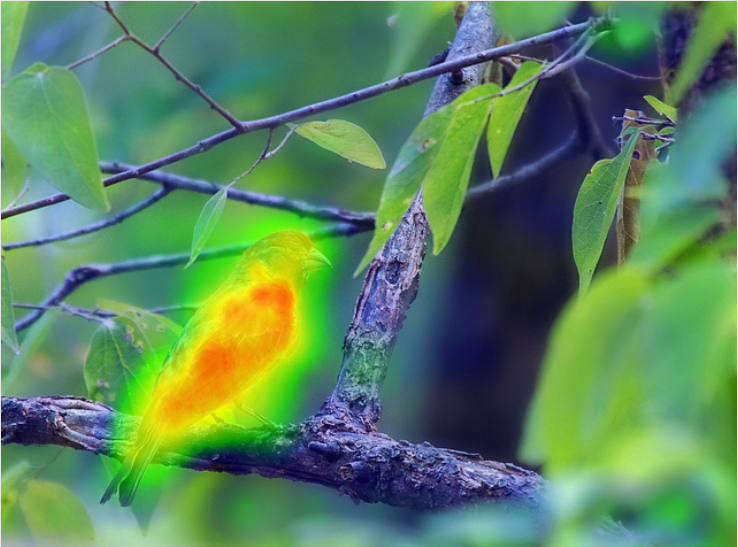}
}
\subfigure[RetinaNet]{
\includegraphics[width=1.4in,height=1.1in]{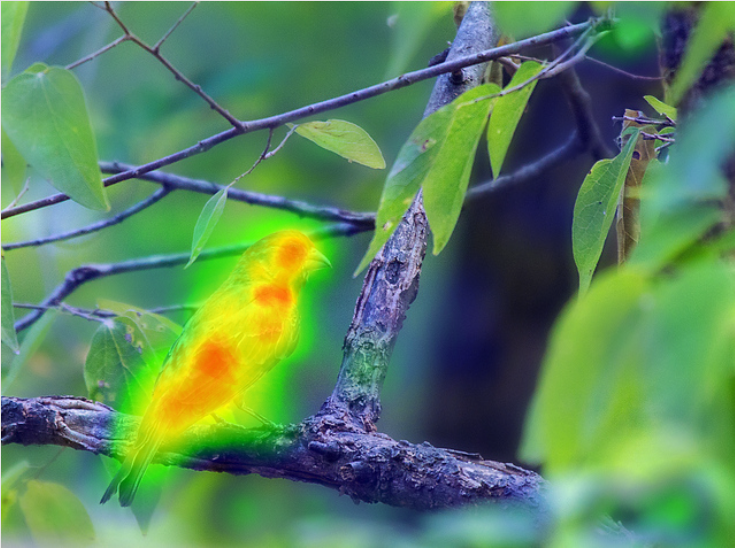}
}
\subfigure[FCOS]{
\includegraphics[width=1.4in,height=1.1in]{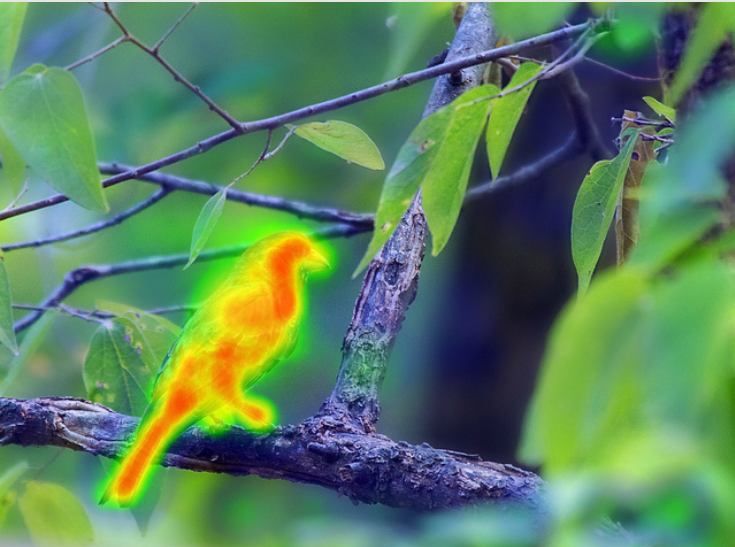}
}
\subfigure[DFMSD (Ours)]{
\includegraphics[width=1.4in,height=1.1in]{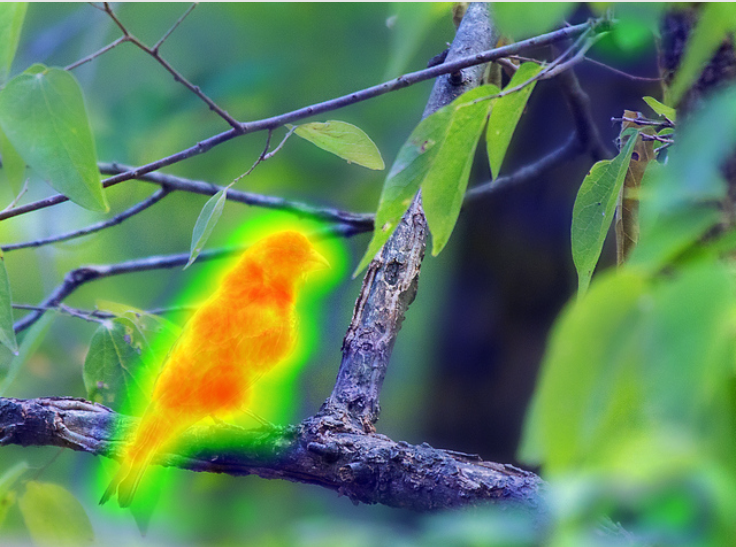}
}
\caption{Comparison of activation maps of heterogeneous detectors evaluated on image ``hummingbird''. Regarding our proposed DFMSD method, we utilize one-stage detector RetinaNet as the student and two-stage detector Faster R-CNN as the teacher. The qualitative results demonstrate significant variances in object-aware perception capability of different detectors characterized by the map intensity. More importantly, our method highlights more object-aware regions with higher intensity and the resulting feature map contains the most discriminative information.}
\label{fig1}
\end{figure}

In terms of detection task, different heterogeneous detectors exhibit significant variances in object perception capability. As can be observed in Fig.~\ref{fig1}, different detectors, including Faster R-CNN~\cite{r10}, RetinaNet~\cite{r8}, and FCOS~\cite{r20} with the same ResNet50 backbone, exhibit substantial differences in activation maps and variations when transformed into feature masks~\cite{r8,r81}. Despite sharing the same backbone architecture, the teacher and student detectors have diverse representation capabilities due to heterogeneous network structures~\cite{r102}. Consequently, heterogeneous detector heads encode different object-aware semantic clues. Directly transferring the knowledge learned from a teacher model to another heterogeneous student model leads to limited performance improvement, suggesting that a huge gap in semantic-aware capability makes it difficult for the student to learn the useful knowledge from the teacher~\cite{r80}. Thus, the reconstructed student features do not improve the model performance.

To address the above-mentioned drawbacks~\cite{r97,r98}, we have proposed a dual feature masking stage-wise distillation framework termed DFMSD for object detection in this study. Following an attention-guided dual feature masking framework, we integrate a stage-wise adaptation learning module into the dual masking framework for addressing heterogeneous distillation. Since it is not beneficial to directly transfer knowledge from the teacher to the student, we perform stage-wise distillation by firstly allowing the student to learn from a ``weaker'' teacher and subsequently adapting the improved student to a ``stronger'' teacher for distillation refinement~\cite{r4,r82}. In this way, the student model can be better adapted to teachers through progressive distillation, which is conducive to bridging the gap between them~\cite{r80}. Furthermore, we embed the masking enhancement strategy into the stage-wise distillation, such that the ``stronger'' teacher in latter distillation stage can benefit from strengthened object-aware masking regions for improved feature-masking reconstruction~\cite{r83}. In addition, we further perform semantic alignment using Pearson correlation coefficients~\cite{r91} to generate consistent teacher-student feature distributions~\cite{r7,r84}. Through the above improvements, we can handle heterogeneous networks within a dual feature-masking distillation framework. Extensive experiments for detection tasks have demonstrated the superiority of our proposed method in both heterogeneous and homogeneous distillation scenarios~\cite{r85}. The contributions of this study can be summarized as follows:                                                     

\begin{itemize}
    \item We have developed a dual feature-masking stage-wise distillation framework (DFMSD) by integrating a stage-wise adaptation learning (SAL) module into the dual masking network for bridging the semantic gap between heterogeneous teacher and student models. It enables the student to firstly learn from a ``weaker'' teacher and refines the adapted student with a ``stronger'' teacher, such that the knowledge can be better transferred to the student with improved adaptability. 
    \item We further introduce a masking enhancement module into our DFMSD, which can adaptively enhance the object-aware masking regions. In terms of the frequency distribution of the semantic regions, adaptive data enhancement strategy is adopted such that the corresponding masking regions can be strengthened for improving masking feature reconstruction.
    \item For better aligning the heterogeneous networks, we further perform semantic alignment between layer-wise features with Pearson correlation coefficients, yielding consistent teacher-student feature distributions.
    \item  Extensive experiments for detection tasks demonstrate the promise of our method in both homogeneous and heterogeneous distillation settings.
\end{itemize}

The remainder of this paper is structured as follows. After reviewing related work in Section \textcolor {blue} {~\ref{sec2}}, we will elaborate on our method in Section \textcolor {blue} {~\ref{sec3}}. In Section \textcolor {blue} {~\ref{sec4}}, we conduct extensive experimental evaluations before the paper is finally concluded in Section \textcolor {blue} {~\ref{sec5}}.

\section{Related Work}\label{sec2}

In this section, we comprehensively review recent advances in object detection and knowledge distillation, both of which are closely related to our approach.

\subsection{Object Detection}\label{subsec21}


It is widely acknowledged that current object detection methods based on deep models can be roughly classified into three categories: anchor-based detectors~\cite{r10,r25}, anchor-free detectors~\cite{r20}, and end-to-end detectors~\cite{r26}. Anchor-based detectors, which consist of two-stage detectors~\cite{r29,r10,r27,r28} and one-stage detectors~\cite{r11,r30,r31}, usually rely on predefined anchor boxes to achieve accurate object detection and localization. In particular, one-stage detectors enjoy a preferable trade-off between efficiency and accuracy by directly classifying and regressing anchors without generating object proposals in advance. Unlike anchor-based detectors, anchor-free approaches including keypoint-based CornerNet~\cite{r65} and center-based CenterNet~\cite{r66,r67}, avoid predefined anchor boxes and can directly predict the object location with desirable flexibility. With the boom of Transformer architecture, recent years have witnessed great success of advanced end-to-end Transformer-based detectors such as DETR~\cite{r26,r24}. They enjoy unparalleled long-range global modeling capability, whereas expensive computational resources and costs are inevitable.

In object detection, there is a huge gap between heavyweight and lightweight detectors. In particular, the heavyweight models, which are in pursuit of high performance, typically require complex backbone structures and significant computational resources~\cite{r24,r33,r34}. Consequently, designing lightweight and efficient detectors with lower complexity and real-time performance is sought-after in practical applications. Since knowledge distillation techniques enable the transfer of stronger representation power from large networks to smaller ones, it facilitates the design of lightweight backbone networks with performance close to that of larger networks~\cite{r81,r87,r88}.

\subsection{Knowledge Distillation}\label{subsec22}


Serving as an effective means of model compression, knowledge distillation maintains the compact structure of lightweight models with significantly improved performance. The earliest work dates back to \cite{r2} where soft labels obtained by a teacher network are incorporated into the loss of a student network, allowing the student network to learn probability distribution consistent with the teacher network for classification. In recent years, dramatic progress has been made in knowledge distillation, and we will comprehensively review different approaches to knowledge distillation.

\subsubsection{Feature-based knowledge distillation}\label{subsec221}

Feature-based distillation methods help the student model mimic the teacher counterpart to generate features with improved representation power. The first feature-based distillation method is known as FitNets in \cite{r5} which demonstrated that semantic information from intermediate layers can also be learned by the student network as implicit knowledge. Hence, distillation techniques have been widely applied to various downstream tasks. Li et al.~\cite{r40} utilized region proposals from the larger network to assist the smaller network in learning higher-level semantic information. Dai et al.~\cite{r12} developed the GID framework which selects specific distillation regions based on differences between student and teacher networks. Yang et al.~\cite{r15} proposed FGD, which separates foreground and background for allowing the student model to learn from regions of interest and global knowledge distilled from the teacher network through simultaneous local and global distillation.


\subsubsection{Masked feature generative distillation}\label{subsec222}

Different from feature distillation techniques, masked feature distillation approaches enable the student model to reconstruct features from selectively masked areas instead of directly learning from the teacher feature. The first masked distillation framework is MGD~\cite{r42} which randomly masks the feature maps of the student model and then reconstructs them from the teacher network. However, random masking may introduce additional noise, leading to biased feature map with impaired representation capability. To identify the importance of the masked areas, attention-driven masked feature distillation methods have been proposed to improve the object-aware perception of the student model. Yang et al.~\cite{r43} proposed an adaptive masking distillation method, termed AMD, for object detection. On the one hand, AMD encodes the importance of specific regions by performing spatially adaptive feature masking, allowing the student model to learn more significant object-aware features from the teacher network. On the other hand, to enhance target perception capabilities, AMD employs a simple yet efficient SE block to generate helpful channel-adaptive cues for the student model. Based on AMD, Yang et al.~\cite{r19} further proposed a dual masking knowledge distillation method, termed DMKD~\cite{r19}. Unlike previous masking-based algorithms, DMKD~\cite{r19} simultaneously focuses on both spatial and channel dimensions, which respectively characterize important spatial regions and channel-wise semantic information. Therefore, it significantly benefits student feature reconstruction and helps to improve the distillation performance, demonstrating superior performance compared to the previous methods. Compared to the aforementioned methods which are essentially one-stage distillation methods, our proposed approach performs stage-wise distillation so that the student can be progressively adapted to multiple teachers in different stages for bridging the gap between heterogeneous networks. To our knowledge, this is the first dual feature-masking stage-wise learning framework for addressing heterogeneous distillation.

\begin{figure*}[ht]
\centering
\includegraphics[width=180mm]{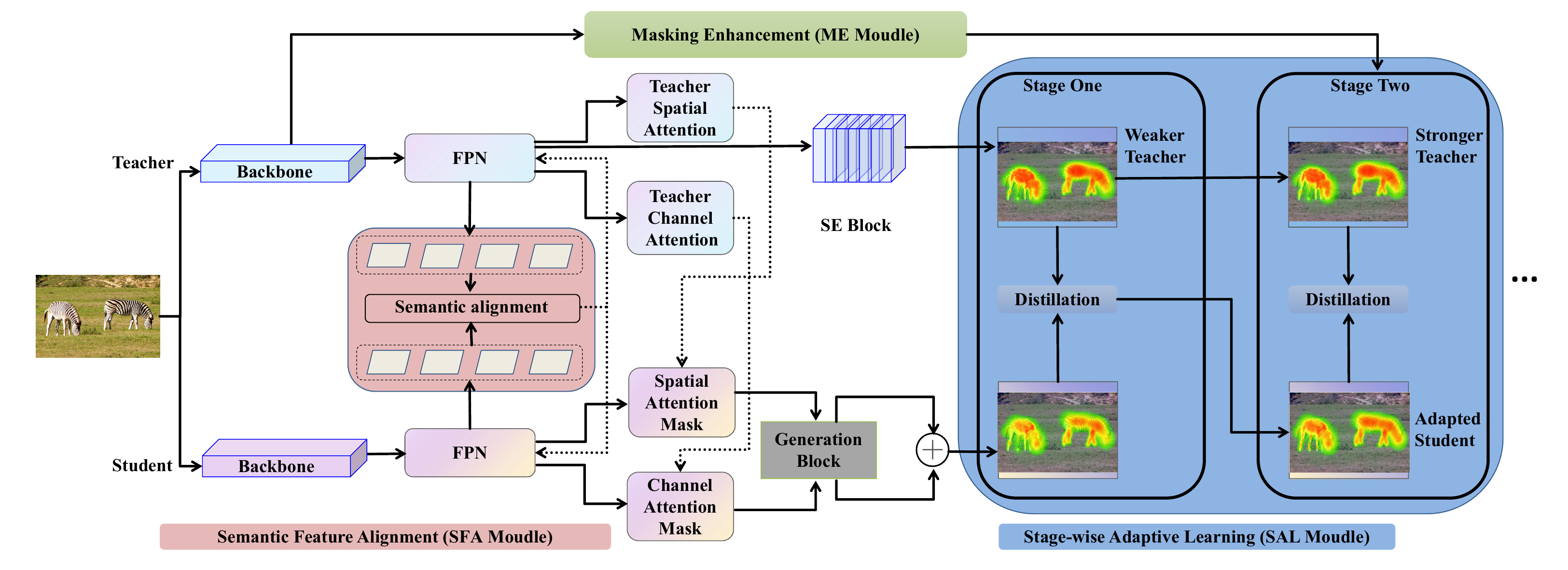}
\caption{Our proposed DFMSD distillation framework. Following the dual-masked knowledge distillation (DMKD) framework where both spatially salient regions and informative channels are identified, a stage-wise adaptive learning strategy (SAL) is integrated, allowing the student network to progressively learning from different heterogeneous teacher networks with improved adaptability. Simultaneously, a masking enhancement module is incorporated into the SAL so that the object-aware masking regions are enhanced for improving masking feature reconstruction. In addition, the semantic feature alignment is performed at each FPN layer between teacher and student backbones, producing consistent feature distribution for further bridging the teacher-student gap.}
\label{fig2}
\end{figure*}


\subsubsection{Heterogeneous Knowledge Distillation}\label{subsec223}

In knowledge distillation, the diversity between the teacher and the student networks poses a great challenge to knowledge transfer and is detrimental to distillation performance, especially when they have heterogeneous network architectures. To address this challenge, MimicDet~\cite{r35} introduced a refinement module that mimics the workflow of two-stage detectors and performs feature alignment between the heads of the teacher and student networks for distillation. G-DetKD~\cite{r36} was the first work to propose a universal distillation framework applicable to object detection. It performs soft matching at all pyramid levels to provide guidance. However, combining different levels of student features by learning similarity scores before feature imitation does not fundamentally bridge the semantic gap. In HEAD~\cite{r1}, an assistant network, which has the same detection head as the teacher detector and learns directly from the teacher, is introduced into the knowledge distillation framework for connecting the teacher-student detectors. Since the assistant and teacher share the same detection head, the semantic feature gap in heterogeneous teacher-student detectors is effectively bridged for better knowledge transfer. Cao et al.~\cite{r37} developed a knowledge distillation method PKD based on the Pearson correlation coefficient~\cite{r91}, which uncovers the linear correlation between teacher and student features. To eliminate the negative effects of amplitude differences between different Feature Pyramid Network (FPN) stages and channels within and between the teacher-student detectors, the feature maps are firstly normalized to have zero mean and unit standard deviation before the mean square error (MSE) loss between these normalized features is minimized. Wang et al.~\cite{r38} proposed an innovative cross-head distillation pipeline termed CrossKD to mitigate the target conflict issue. This method transfers intermediate features of the student network to the detection head of the teacher network, thereby generating cross-head predictions. Then, knowledge distillation is performed between these newly generated cross-head predictions and the original predictions generated from the teacher model. It guarantees that the KD loss does not influence the weight updates in the detection head of the student network, avoiding conflicts between the original detection loss and the KD loss. In addition, since both cross-head predictions and teacher predictions are generated from sharing parts of the detection head in the teacher network, the cross-head predictions are relatively consistent with the predictions obtained by the teacher. This significantly reduces the discrepancies between the teacher and student detectors, enhancing the stability of training during prediction imitation~\cite{r99,r100}. While these methods can achieve successful heterogeneous distillation, they do not explore feature masking in an adaptive stage-wise distillation manner, leading to the student features with limited boost in representation power. Consequently, the student still far lags behind the teacher and the large gap still exists between the heterogeneous networks. In contrast, our method addresses heterogeneous distillation by performing dual feature-masking stage-wise learning, thereby steadily improving the student feature and effectively reducing the gap between heterogeneous networks.

\section{Proposed Method}\label{sec3}

Since our proposed method essentially falls into the category of masked feature distillation, we will firstly introduce the formulation of feature distillation. Based on the feature distillation formulation, we will present an attention-directed dual masking distillation framework followed by our method. Furthermore, we will elaborate on our proposed dual feature masking stage-wise distillation (DFMSD) framework with three crucial components.

\subsection{Problem Formulation}

Feature distillation allows feature-level knowledge transfer from the teacher model to the student model to generate sufficiently descriptive features that are competitive with the teacher counterpart. Mathematically, it can be achieved with the following distillation loss function:

\begin{equation}
L_{Fea} = {\textstyle \sum_{l=1}^{L}} \frac{1}{N_{l} } \sum_{c}^{C} \sum_{h}^{H} \sum_{w}^{W} \left \|F_{c,h,w}^{T} -\Phi (F_{c,h,w}^{S} )  \right \|_{2}^{2}  
\end{equation}
where $L$ denotes the number of layers in the FPN after the backbone networks, $N_l$ represents the feature size of $l$-th layer, while $C$, $H$, and $W$ indicate the number of channels, height, and width of the feature map, respectively. $F^{T}$ and $F^{S}$ denote respective features generated from the teacher and the student model. $\Phi(\cdot)$ indicates the linear projection layer, which is capable of aligning $F^{S}$ with $F^{T}$ in the feature resolution.

Recent studies have suggested that learning and reconstructing student features from the teacher model are considered to be a preferable alternative to feature imitation in the conventional feature distillation paradigm~\cite{r42,r43}. More specifically, expressive features can be reconstructed from selectively masking regions on the feature maps of the student network, which is also known as masked feature distillation. In particular, attention-directed masked feature distillation has improved the prototype masked generative distillation framework in which masked regions are randomly generated~\cite{r42}. Recently, a dual masking knowledge distillation framework, termed DMKD~\cite{r19}, is proposed to comprehensively encode object-aware semantics into the student network. More specifically, the dual attention maps derived from the teacher networks to capture both spatially important and informative channel-wise clues are formulated as:
\begin{equation}
A^{c} =Sigmoid\left ( \frac{1}{HW\tau }\sum_{h=1}^{H} \sum_{w=1}^{W} \left \langle  F_{h,w,1}^{T}   ,..., F_{h,w,C}^{T}   \right \rangle  \right )
\end{equation}

\begin{equation}
A^{s} =\phi _{align}\left (   Sigmoid\left ( \frac{1}{C\tau }\left \langle \left \| F_{1}^{T}  \right \|_{2}^{2} ,...,\left \| F_{n}^{T}  \right \|_{2}^{2}  \right \rangle  \right )\right )
\end{equation}
where $A^{c}$ $\in$ $R^{C\times1\times1}$ and $A^{s}$ $\in$ $R^{1\times H \times W}$ represent the channel and spatial attention maps, respectively. Then, attention-guided feature masking is performed before improved masked feature reconstruction is achieved via SE and generation modules~\cite{r19}.

\subsection{Our DFMSD Framework}\label{subsec31}


Although the above-mentioned dual-masked feature distillation scheme is capable of reconstructing student features with improved representation power, it fails to transfer knowledge well from a teacher model to a student model when they have diverse network architectures, thereby achieving deteriorating performance for the heterogeneous distillation task. To alleviate this problem, we propose a dual feature-masking stage-wise knowledge distillation method for object detection, termed DFMSD, in this study. Fig.~\ref{fig2} illustrates the framework of our proposed DFMSD model. Built on DMKD, the stage-wise adaptive learning strategy is integrated into the dual-masked distillation framework to progressively adapt the student to the teacher in separate stages, which contributes to bridging the gap between heterogeneous networks. Meanwhile, a masking enhancement module is also introduced to adaptively enhance object-aware masking regions according to the frequency distribution characteristics, so that stage-wise distillation is further improved with enhanced feature masking. In addition, semantic alignment is performed between teacher-student FPNs via Pearson Correlation Coefficient~\cite{r91} for generating consistent feature distributions. Thus, our DFMSD network is capable of narrowing the teacher-student discrepancy with improved heterogeneous distillation performance. Next, we will elaborate on the three critical components mentioned above within our DFMSD network.

\begin{figure}[ht]
\centering
\includegraphics[width=88mm]{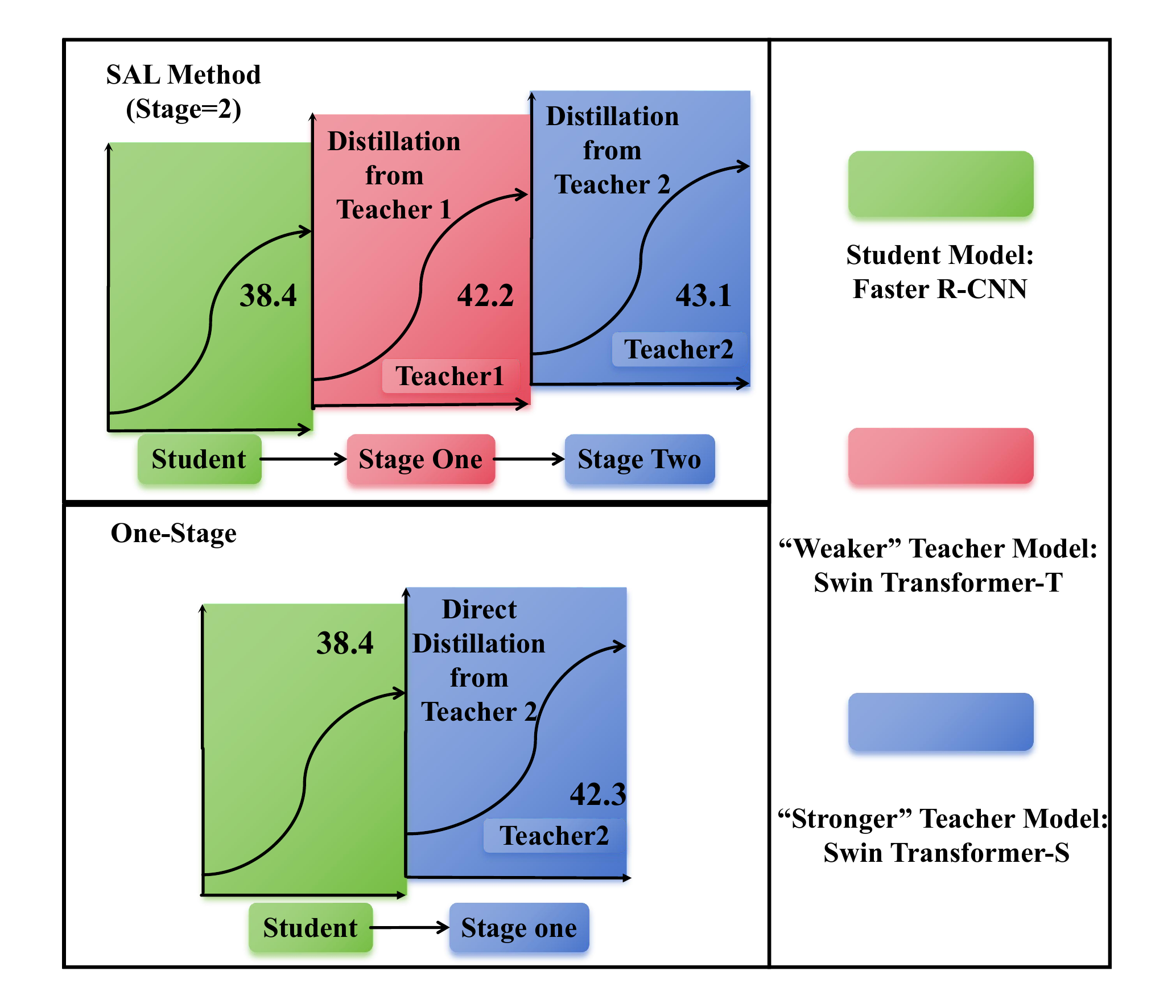}
\caption{Illustration of our SAL mechanism for adaptively improving the distillation performance. With Swin Transformer~\cite{r24} and Faster R-CNN~\cite{r10} used as respective teacher and the student networks, two-stage SAL mechanism firstly improves the Faster R-CNN detector from 38.4\% to 42.2\% with a Swin-Transformer-T model~\cite{r24}, and then further boosts the student performance to 42.9\% with a more powerful Swin-Transformer-S detector~\cite{r24}. In contrast, the conventional one-stage distillation approach only improves the Faster R-CNN model to 42.3\% accuracy which is roughly the first-stage distillation performance within SAL.}
\label{fig3}
\end{figure}

\begin{figure*}[h]
\centering
\subfigure[Teacher]{
\includegraphics[width=1.7in,height=1.2in]{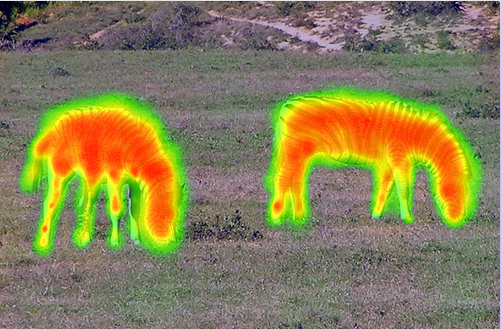}
}
\subfigure[Student]{
\includegraphics[width=1.7in,height=1.2in]{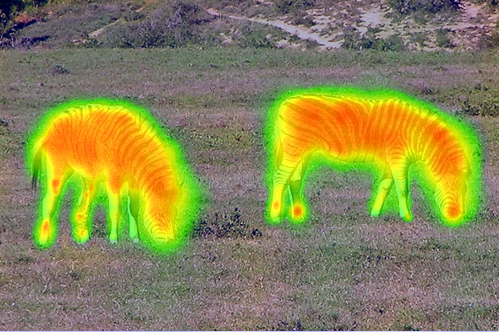}
}
\subfigure[After one-stage distillation]{
\includegraphics[width=1.7in,height=1.2in]{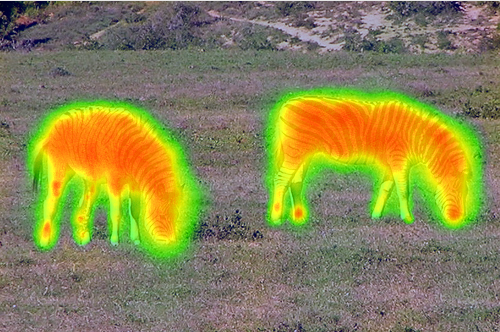}
}
\subfigure[After two-stage distillation]{
\includegraphics[width=1.7in,height=1.2in]{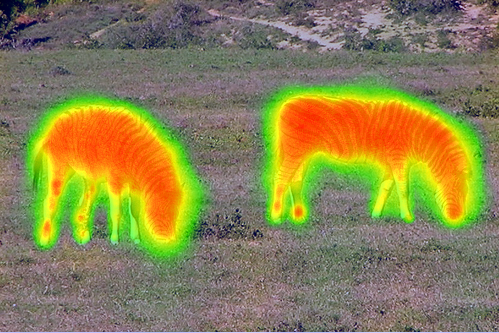}
}
\caption{Comparison of the student feature maps in different distillation stages of our SAL mechanism. (a) and (b) demonstrate two feature maps obtained by the original teacher and student models, respectively. (c) and (d) illustrate the student feature maps generated after the first-stage and the second-stage distillation respectively. It can be observed that distinct object-aware regions can be captured after consecutive distillation stages, yielding sufficiently discriminative feature maps close to the teacher counterparts.}
\label{fig4}
\end{figure*}


\subsection{Stage-wise Adaptive Learning Module}\label{subsec32}

The conventional masked distillation paradigm adopts a one-stage knowledge transfer strategy, in which a student model directly learns from one teacher model via single one-stage learning. However, this ``one-stage learning'' usually makes it difficult for the student model with limited capacity to learn sufficiently from a highly complex teacher model, let alone a heterogeneous teacher model with an entirely different network structure. To narrow the gap between heterogeneous teacher and student networks, we have integrated the stage-wise adaptive learning (SAL) mechanism into the dual masked distillation framework for improving the adaptability of the student model. Different from the previous methods in which only one teacher model is used in the distillation process, our strategy takes advantage of several advanced detectors and allows the student network to adaptively learn from the teachers in separate stages. More specifically, the student model can initially learn from relatively weaker teacher networks in the preceding stages, yielding suboptimal results. Subsequently, the adapted student is utilized as a new student to learn from a stronger teacher network in the latter stages, facilitating a more complete knowledge transfer. With the help of this SAL mechanism, the student network can be better adapted to the teacher model with the progressive distillation stages, and thus the gap between heterogeneous networks can be dramatically bridged. 


The beneficial effects of our SAL module can be illustrated in Fig.~\ref{fig3}. It can be observed that the SAL module significantly benefits improving the distillation performance when heterogeneous Swin Transformer~\cite{r24} and Faster R-CNN~\cite{r10} are respectively used as the teacher and the student detectors. To be specific, the two-stage adaptive learning allows the Swin-Transformer-T~\cite{r24} to boost the performance of the Faster R-CNN model from 38.4\% to 42.2\%, and further improves 0.9\% with a stronger Swin-Transformer-S teacher detector~\cite{r24}, achieving 43.1\% mAP accuracy. This surpasses the traditional one-stage distillation method in which Faster R-CNN directly learns from the Swin-Transformer-S model and reports suboptimal 42.3\% accuracy, which is only on par with the first-stage distillation performance within our SAL.

Fig.~\ref{fig4} intuitively compares the feature maps generated from the student network in different stages using SAL strategy. It can be clearly observed that the student network can capture more object-aware semantic regions after consecutive distillation stages. For example, compared with the original feature map of the student network, more semantically important regions corresponding to the zebras' heads and necks can be uncovered after the first-stage distillation. When the second-stage distillation is completed, the zebra-specific regions can be comprehensively characterized by discriminative feature maps close to the teacher counterparts and readily distinguished from the background regions. This fully suggests that our SAL module not only progressively improves the representation power of the student model but also significantly bridges the gap between heterogeneous teacher and student networks.

\begin{figure}[h]
\centering
\subfigure[High-frequency object]{
\includegraphics[width=1.65in,height=1.1in]{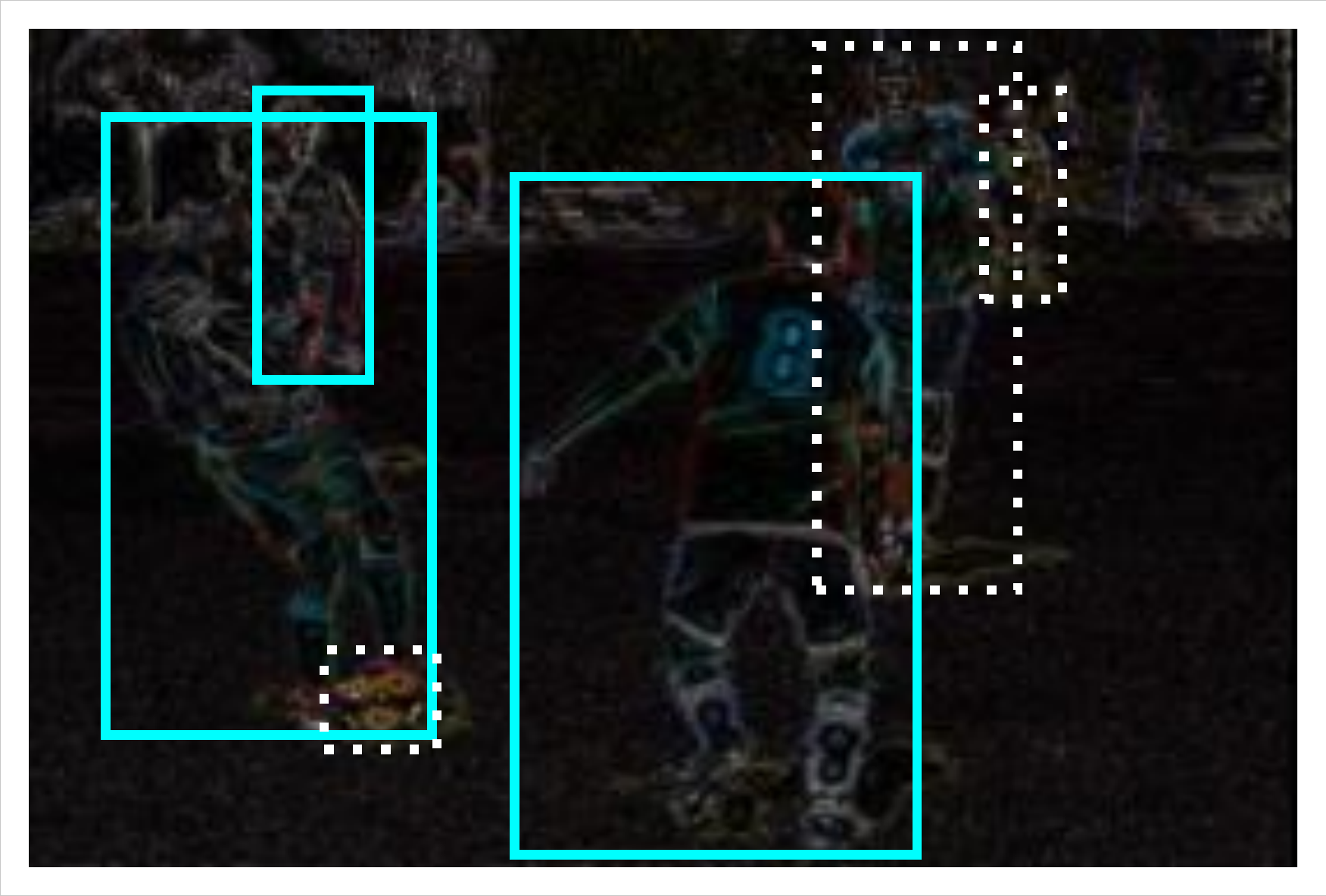}
}
\subfigure[Low-frequency object]{
\includegraphics[width=1.65in,height=1.1in]{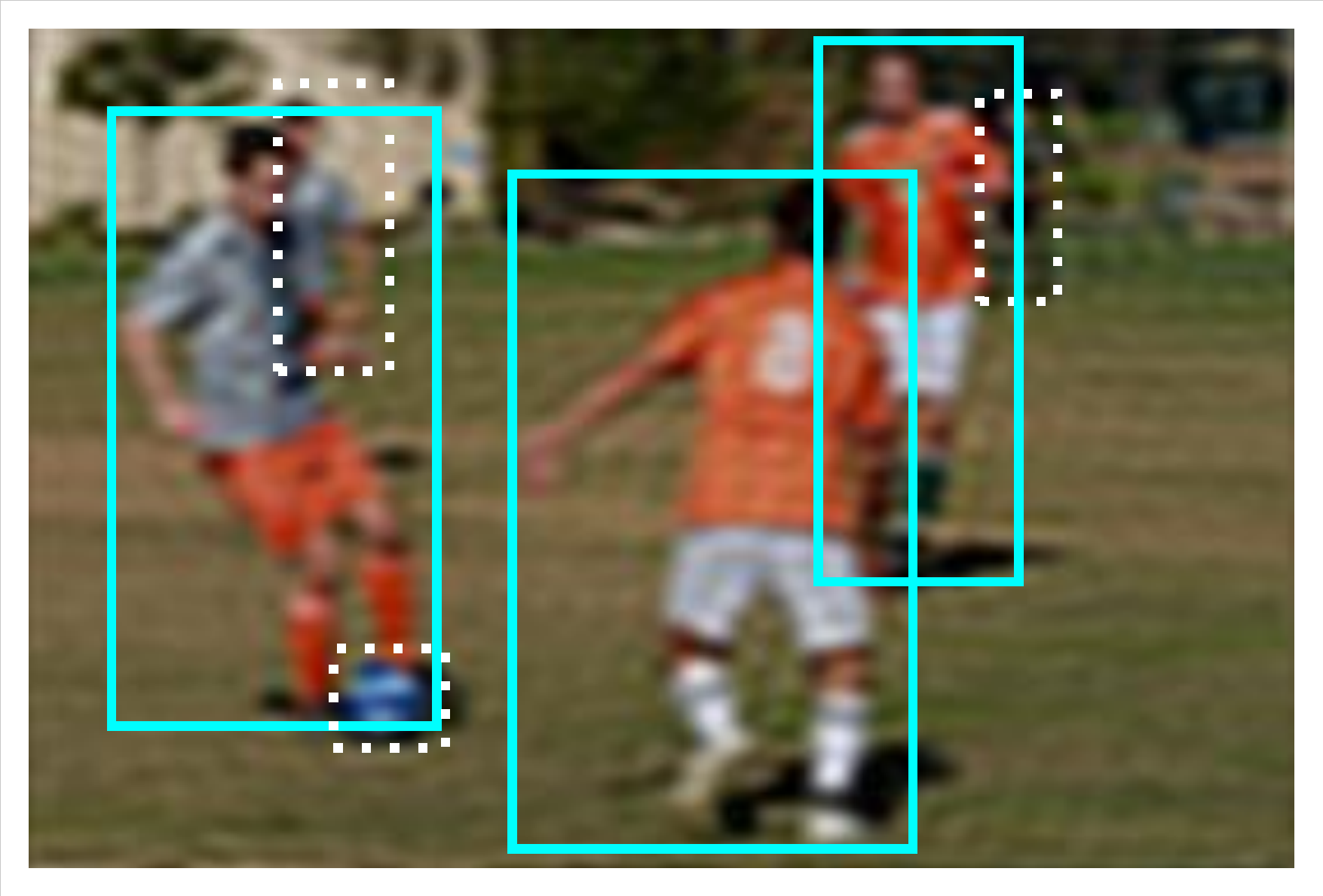}
}
\subfigure[High-frequency masking Map]{
\includegraphics[width=1.65in,height=1.1in]{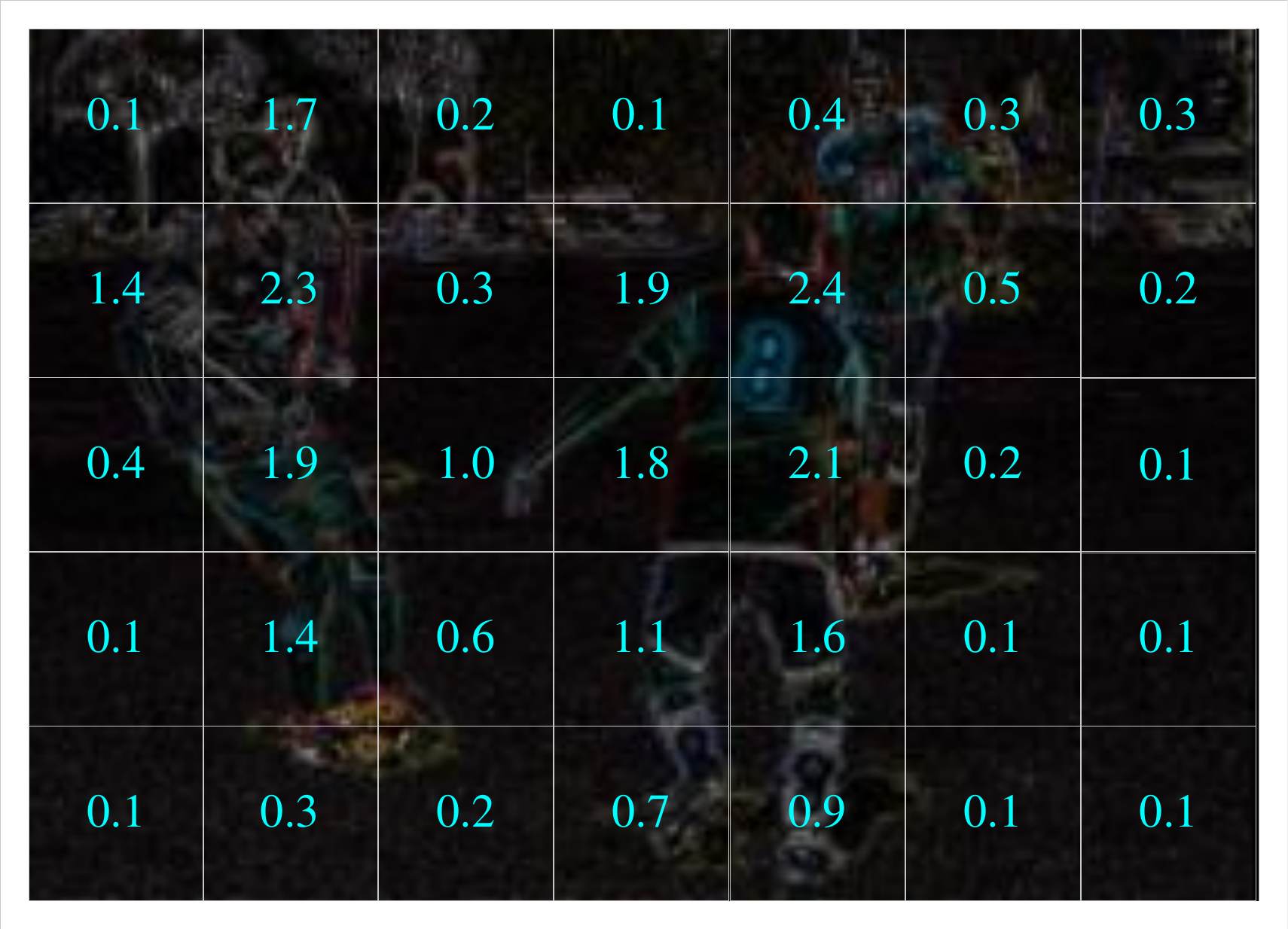}
}
\subfigure[Low-frequency masking Map]{
\includegraphics[width=1.65in,height=1.1in]{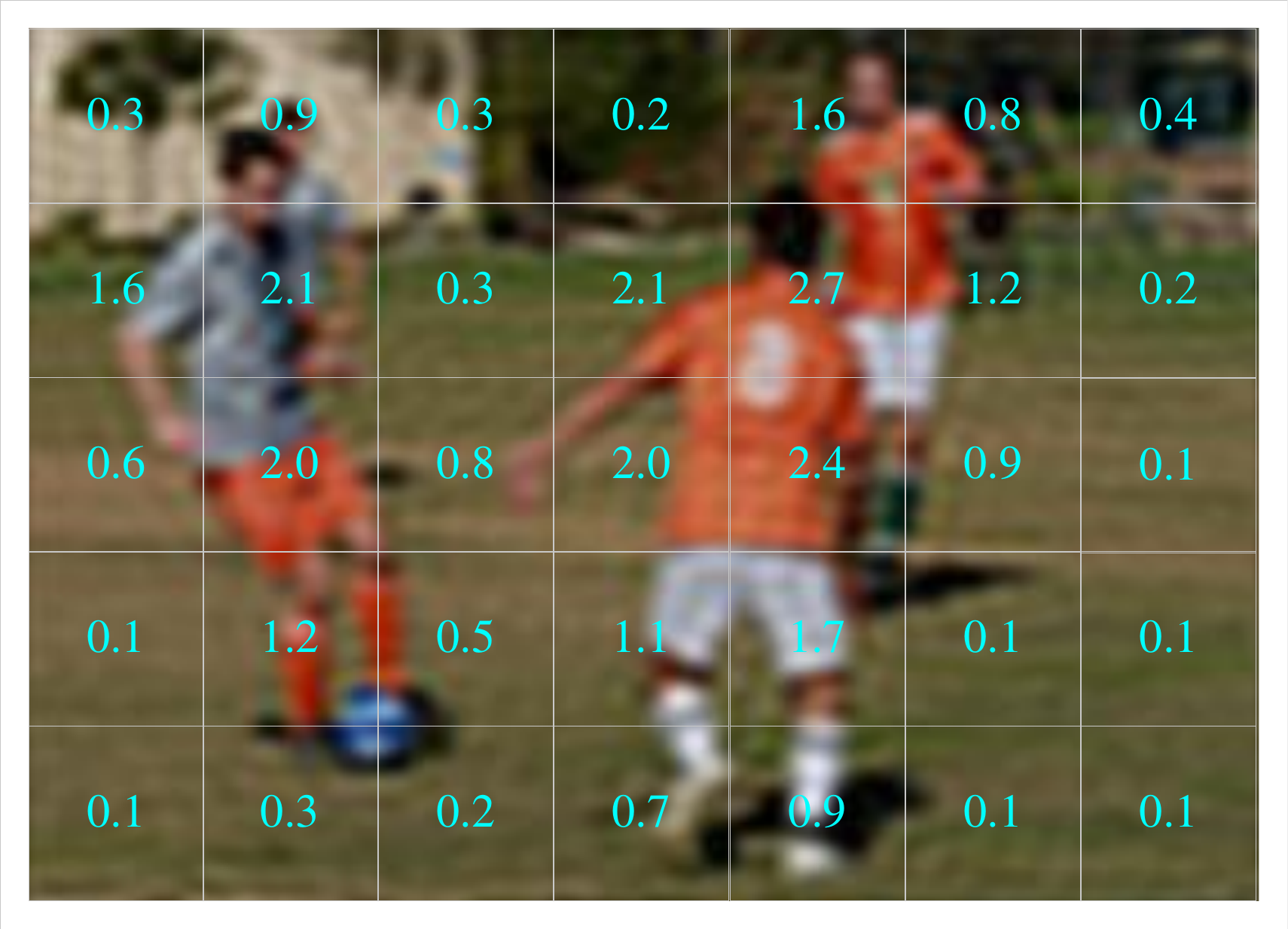}
}
\subfigure[Original masking map]{
\includegraphics[width=1.65in,height=1.1in]{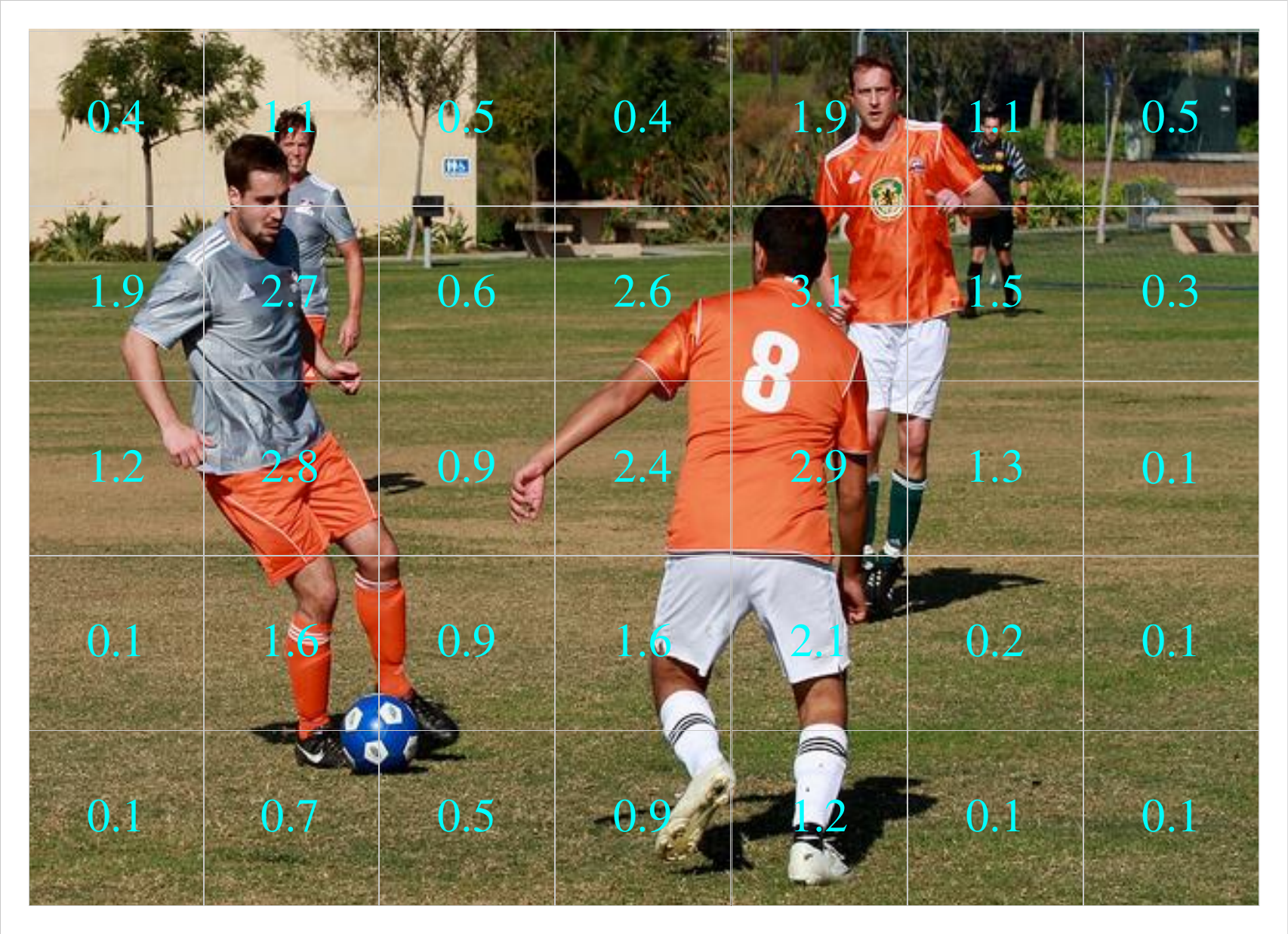}
}
\subfigure[Enhanced masking map]{
\includegraphics[width=1.65in,height=1.1in]{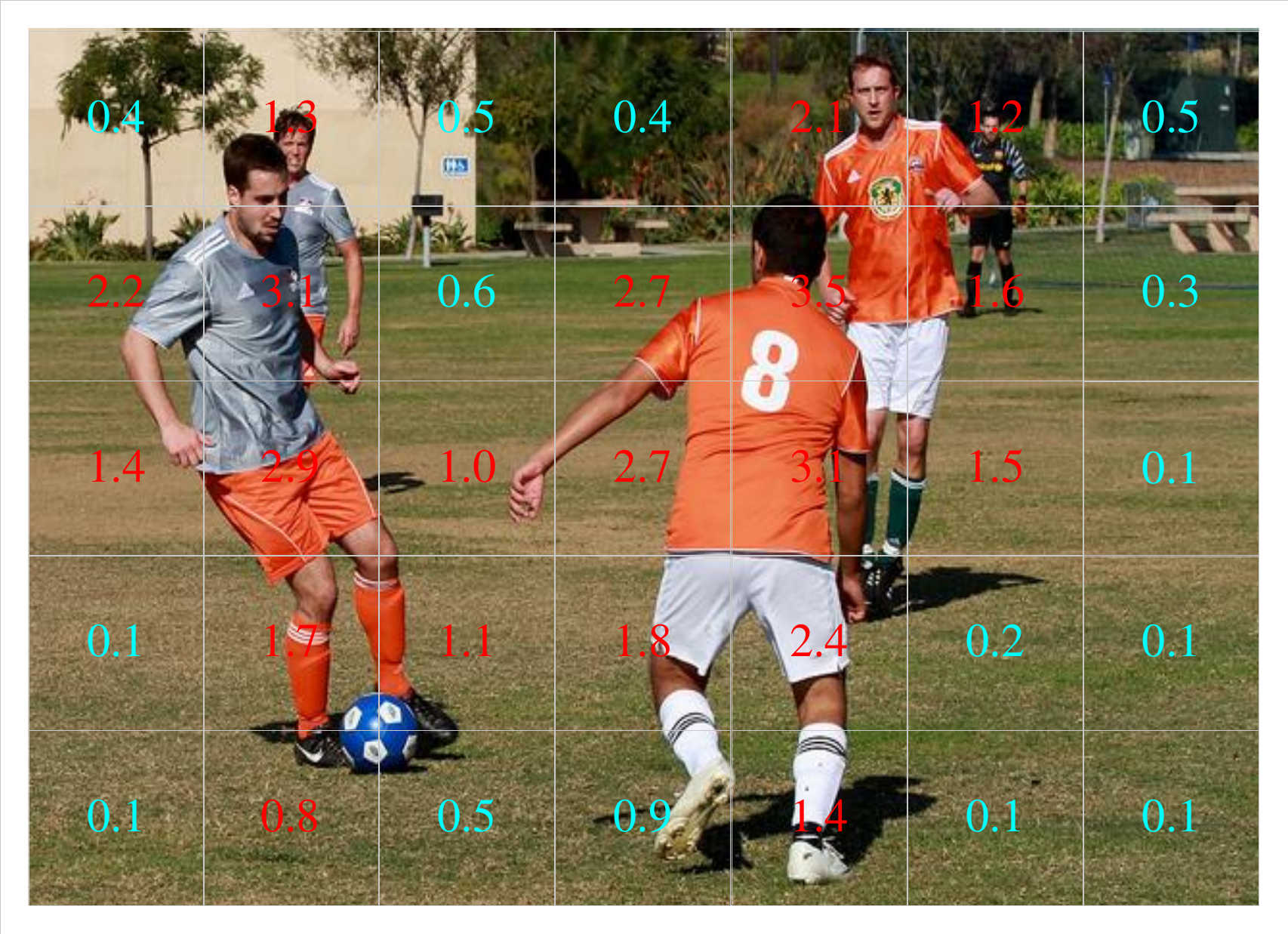}
}
\caption{Comparison of object-aware candidate boxes and region-aware attention score distribution in different frequency domains achieved by RetinaNet detector. It can be clearly observed that the RetinaNet misses some small objects including the football and the far-end partially occluded referee in black even in the high-frequency domain of the image. This can also be demonstrated in the region-specific attention maps where the corresponding regions are low-scored. However, with the help of our adaptive augmentation strategy, the importance of the high-frequency regions corresponding to the small objects is promoted with increased attention scores highlighted in (f), which is beneficial for the subsequent feature masking and reconstruction.}
\label{fig5}
\end{figure}

\begin{figure}
\centering
\subfigure[Original]{
\includegraphics[width=1.4in,height=1.1in]{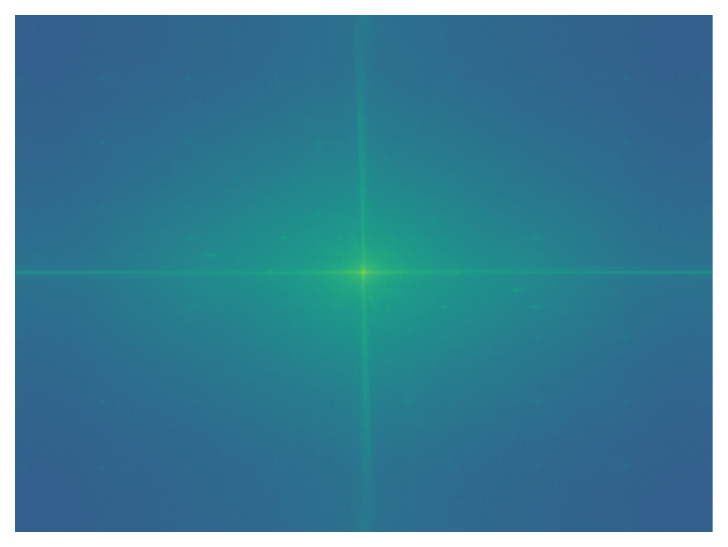}
}
\subfigure[Flipping]{
\includegraphics[width=1.4in,height=1.1in]{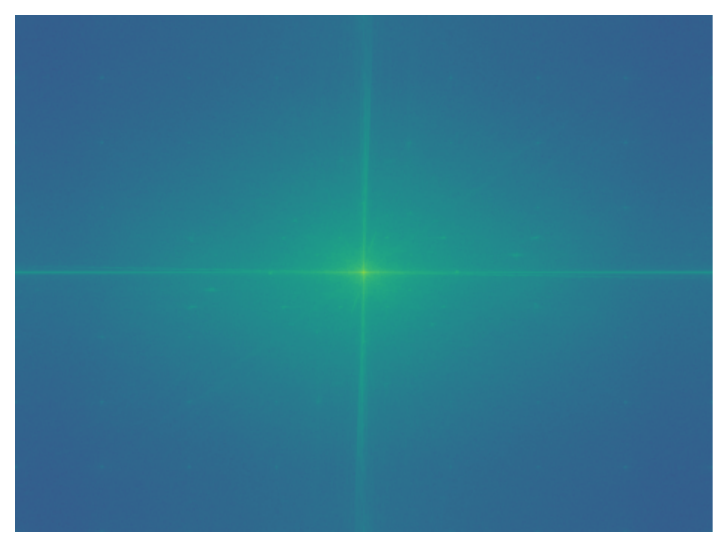}
}
\subfigure[Adding Gaussian noise]{
\includegraphics[width=1.4in,height=1.1in]{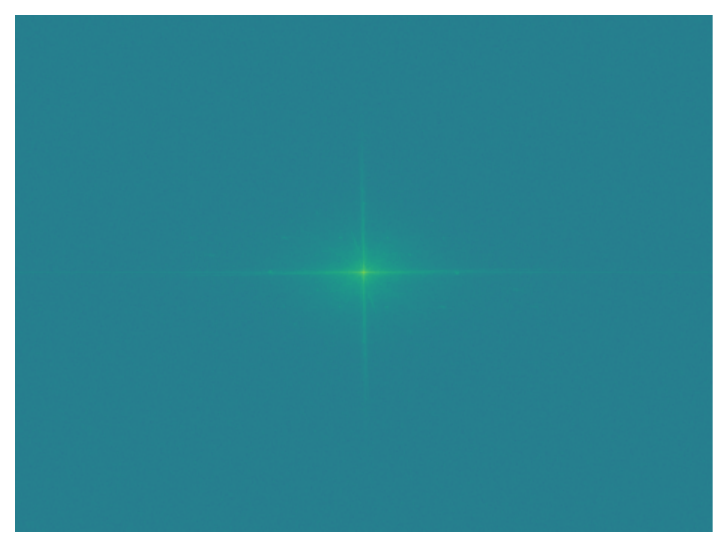}
}
\subfigure[Cropping]{
\includegraphics[width=1.4in,height=1.1in]{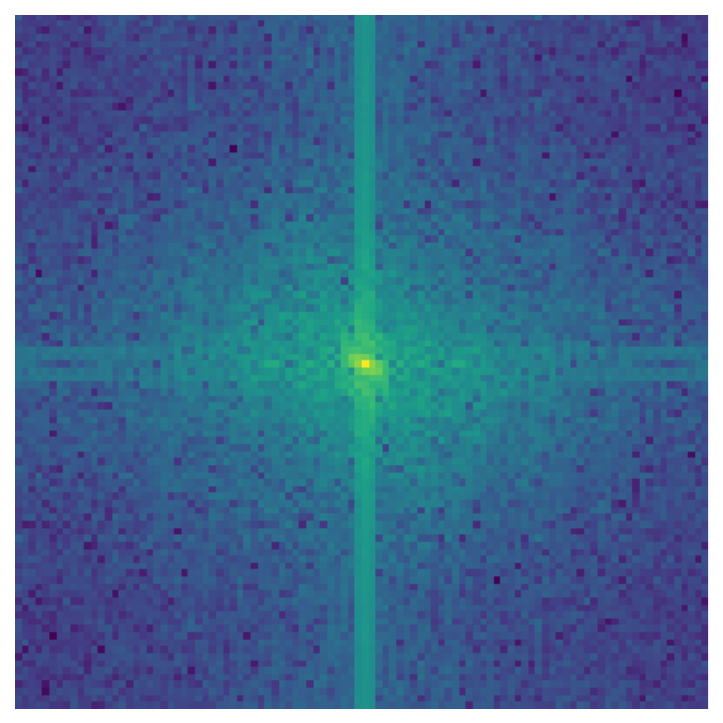}
}
\caption{Comparison of two-dimensional Fourier spectrums of the original image (a), and its variants transformed by flipping (b), adding Gaussian noise (c), and cropping (d). It demonstrates that different augmentation schemes can strengthen specific frequency information. In particular, cropping helps to enhance the low-frequency information while Gaussian noise perturbation can suppress it for better identifying high-frequency smaller objects.}
\label{fig6}
\end{figure}

\subsection{Masking Enhancement module}\label{subsec33}

Prior research explores scale-aware object perception capability of CNN-based models from a frequency perspective~\cite{r45}. It demonstrates that the same detector exhibits diverse detection performance in different frequency domains. More specifically, a CNN-based detector is likely to successfully identify larger objects while missing smaller ones in the low-frequency domain of an image, and vice versa in the high-frequency domain. Thus, when performing attention-directed feature masking on both frequency domains, different attention maps are obtained depending on the variance in object-aware frequency distribution. To be specific, the masked regions of smaller objects corresponding to the high-frequency components are endowed with higher attention scores, while the low-frequency masked regions are usually downplayed in the high-frequency domain. Conversely, low-frequency masked regions corresponding to larger objects tend to receive more attention and outweigh high-frequency regions in the low-frequency domain. However, within our SAL module, a ``weaker'' teacher with limited object-aware capability fails to generate accurate attention maps encoding spatial importance, especially when object-specific frequency distribution in an image is diverse. For example, as shown in Fig.~\ref{fig5}, the RetinaNet detector generates a low attention score in some high-frequency regions corresponding to smaller objects in the high-frequency domain of the image. The low-scored regions are not identifiable for feature masking, which is detrimental to accurate detection of smaller objects, including the football and the far-end partially occluded referee in black.


To further benefit subsequent distillation, we have introduced a masking enhancement module into our SAL module to improve object-aware perception capability. In terms of our masking enhancement strategy, data augmentation methods are adaptively applied to an image according to its object-specific frequency distribution, generating enhanced masking regions for feature reconstruction. For example, a proper augmentation method should strengthen the high-frequency information in an image dominated by small objects, such that more regions corresponding to the high-frequency small objects are identified as semantically important for feature masking. In contrast, when most objects in an image are medium-size or large-size, more low-frequency regions should be enhanced by adaptive data augmentation scheme to identify larger objects in the image.

To investigate the frequency attributes of different data augmentation~\cite{r93,r94} methods, including random flipping~\cite{r48}, random cropping~\cite{r49}, and Gaussian noise perturbation~\cite{r50}, we have performed detailed analyzes to explore the effects of various augmentation approaches on the original images in the frequency domain. More specifically, we performed two-dimensional Discrete Fourier Transform (DFT)~\cite{r92} on images including an original unaltered image and its variants processed with different data augmentation methods, yielding a variety of Fourier spectrums used to intuitively demonstrate the frequency characteristics of different augmentation methods. As shown in Fig.~\ref{fig6}, flipping the image produces a Fourier spectrum that resembles the original one without essentially changing its attribute characteristics. When adding Gaussian noise to the image, however, it can be observed that close-to-center frequency amplitude is suppressed in the frequency spectrum, which implies that Gaussian noise perturbation could benefit uncovering high-frequency small objects in the image. In contrast, images subjected to random cropping exhibit higher amplitude in the close-to-center region of the Fourier spectrum, suggesting that the low-frequency information of the image is strengthened.

Since different augmentation strategies can boost specific frequency information, we attempt to perform an adaptive data augmentation technique on an image according to its object-aware frequency characteristics, such that the corresponding masking regions can be enhanced for feature reconstruction with improved representation power. On the one hand, we adopt a cropping augmentation approach to enhance the low-frequency components in an image hardly containing small object. To be specific, a randomly proportional cropping strategy is employed to adjust the edges of the image, which not only enhances the low-frequency clues of the image, thereby allowing the model to accurately identifying and localizing large-object regions. On the other hand, we add high-frequency Gaussian noise to an image predominantly featuring small objects for enhancing the high-frequency information. Specifically, high-frequency noise is sampled from a normal distribution with a mean of 0 and a variance of $\sigma$², denoted as $\mathcal{N}$(0, $\sigma$²), and added to the original clean image with a certain probability. In this way, we can enhance the high-frequency object-aware regions of the images while maintaining primary feature information, thereby helping the detector to capture small objects more accurately. The resulting adaptively augmented data are delivered to the ``stronger'' teacher detector in the last stage of our SAL module for generating enhanced attention masks.

\begin{figure*}[ht]
\centering
\includegraphics[width=180mm]{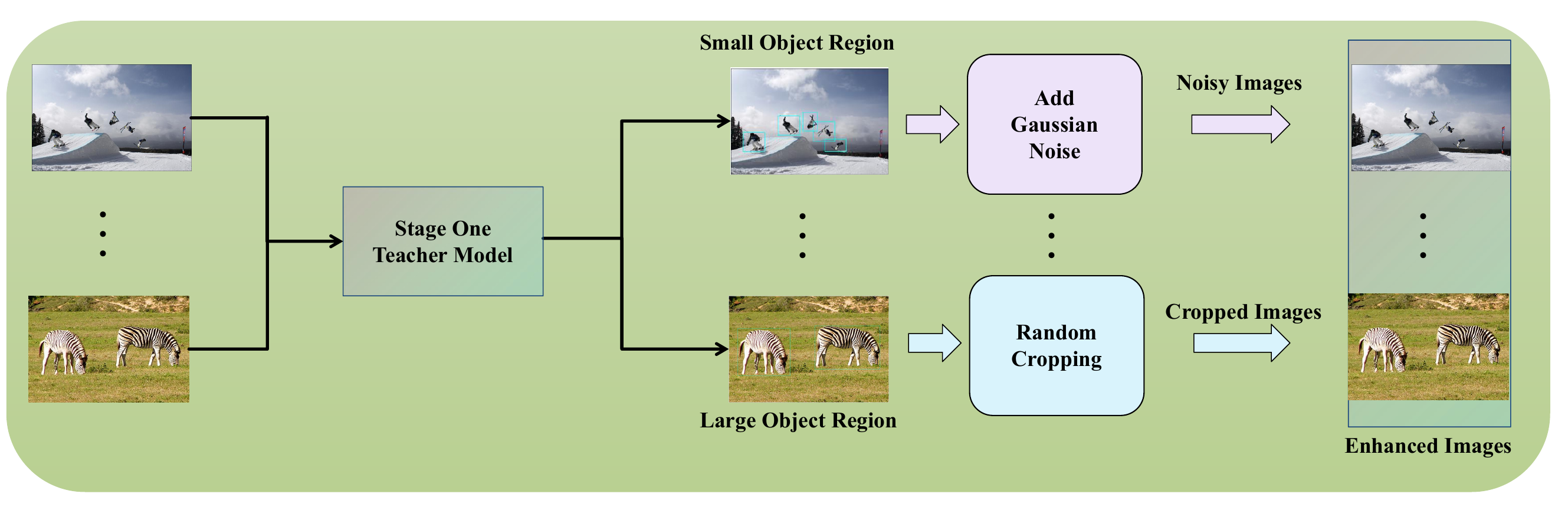}
\caption{The pipeline of our adaptive data augmentation method. For images with various object distributions, our method can adaptively perform data augmentation to enhance object-aware frequency information. Thus, enhanced object-aware attention masks can be obtained by the ``stronger'' teacher within our SAL module. }
\label{fig7}
\end{figure*}

Fig.~\ref{fig7} demonstrates our introduced data augmentation strategy. For images with different object-aware distributions used as input, candidate object regions can be derived from the ``weaker'' teacher model in the previous stage. Then, adaptive augmentation approaches are employed depending on the object-specific frequency characteristics.

Mathematically, our proposed feature masking adaptive data augmentation method can be formulated as:
\begin{equation}\label{eq4}
K_{mask}^{size} =
\begin{cases}
k^{big}(x)  & \text{ if } Area(x)\ge \lambda  \\
k^{small}(x)  & \text{ if } Area(x)<\lambda
\end{cases}
\end{equation}
where $Area(x)$ represents the summed area of all the candidate bounding boxes in image $x$ derived from the teacher detector in the first distillation stage. $\lambda$ denotes the predefined threshold that can help to distinguish whether an image predominantly contains relatively smaller or larger object-aware regions. When an image $x$ predominantly constitutes relatively smaller objects indicated as $Area(x)<\lambda$, Gaussian noise is added to the image for enhancing the high-frequency masking regions corresponding to smaller objects. In contrast, low-frequency object-aware masking regions can be enhanced via cropping mechanism such that the larger-object masking regions receive more attention. Thus, adaptive enhanced masking regions can be obtained for improved feature reconstruction.

Following~\cite{r58, r89}, in addition, adversarial examples are introduced for further mining inconsistent knowledge within the teacher model, which is conducive to improving the semantic perception capability of the student network~\cite{r95,r96}.

\begin{figure*}[h]
\centering
\subfigure[Teacher-student feature distribution (before alignment)]{
\includegraphics[width=3.4in,height=2.0in]{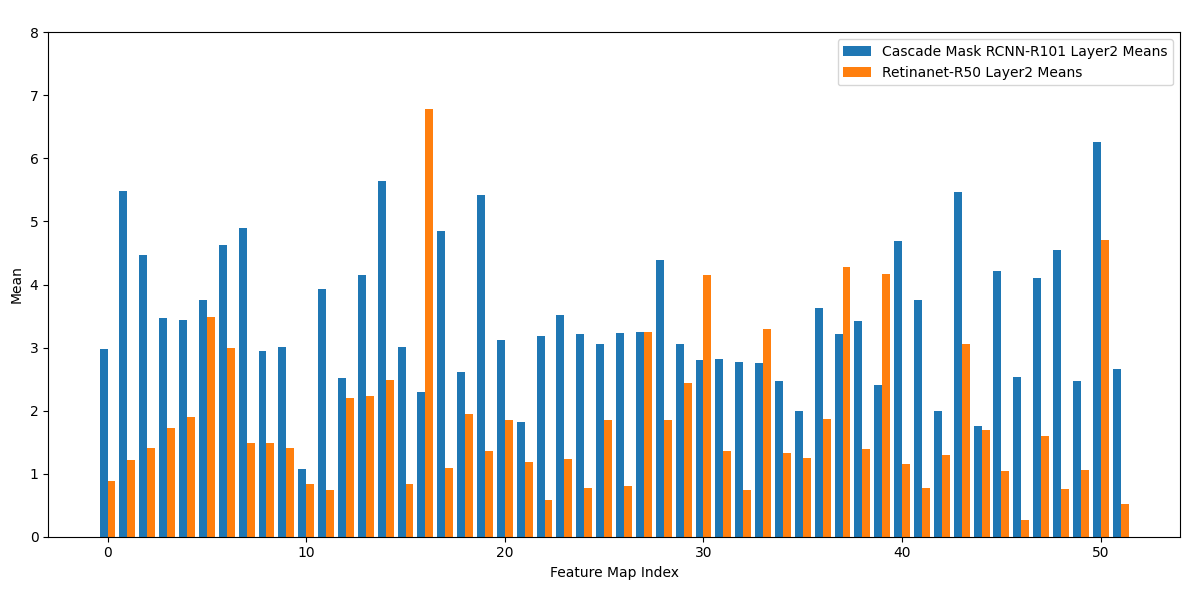}
}
\subfigure[Teacher-student feature distribution (after alignment)]{
\includegraphics[width=3.4in,height=2.0in]{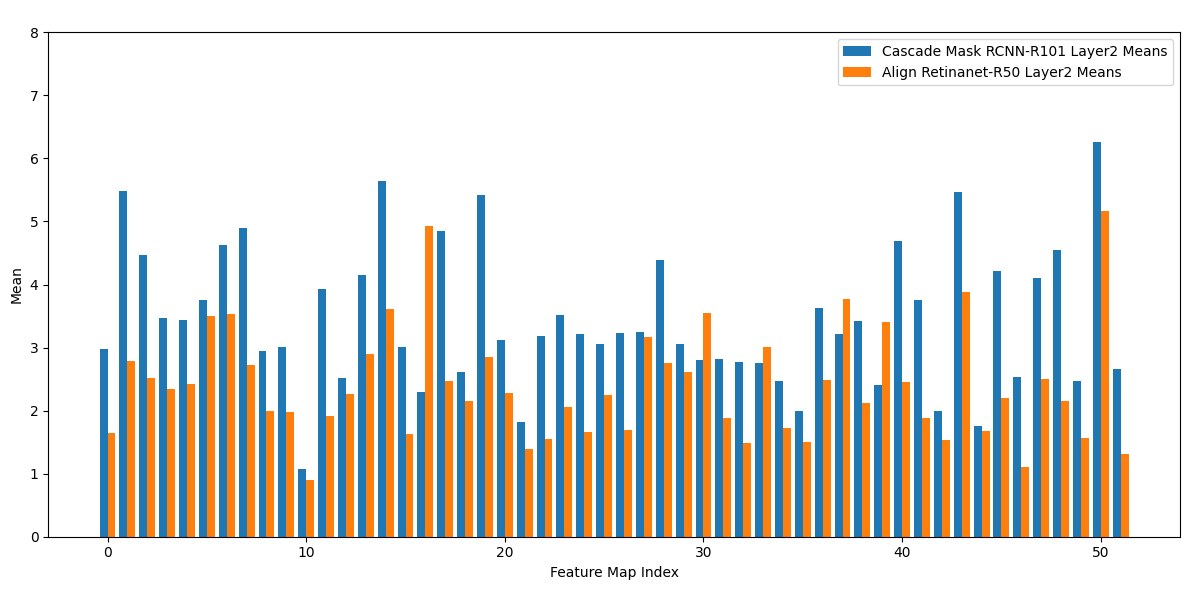}
}
\subfigure[Student feature map]{
\includegraphics[width=2.2in,height=1.6in]{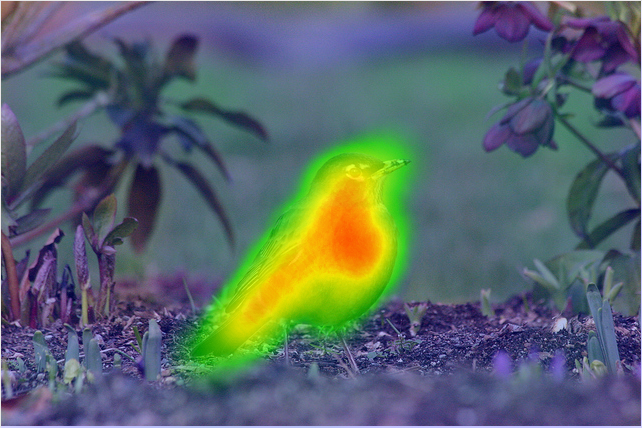}
}
\subfigure[Teacher feature map]{
\includegraphics[width=2.2in,height=1.6in]{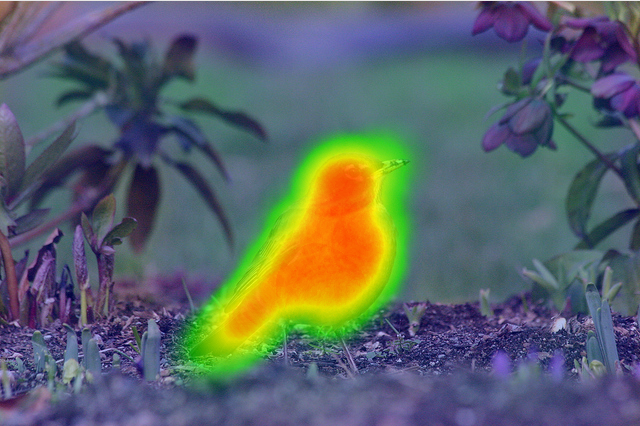}
}
\subfigure[Student feature map (after alignment)]{
\includegraphics[width=2.2in,height=1.6in]{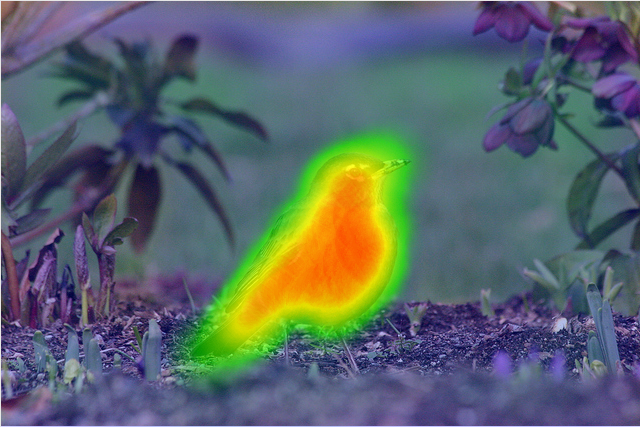}
}
\caption{By aligning the teacher and student models to the same dimensionality, we computed the mean feature values for each channel, grouped all channels into several channel groups, computed the mean again for each channel group, and finally plotted the bar charts showing the feature distribution of both the teacher and the student models before alignment (a) and after alignment (b). The feature maps of the student, the teacher, and the aligned student at the second layer of the FPN (P2) are displayed from (c) to (e). It can be observed that the object perception capability of the aligned student network is significantly enhanced after alignment.}
\label{fig8}
\end{figure*}

\subsection{Semantic Feature Alignment Module}\label{subsec34}

Due to the teacher-student gap, there is also a significant variance in the feature semantic awareness at each FPN level between heterogeneous networks. As shown in Fig.~\ref{fig8}, there is a significant disparity between the feature distributions of the student and teacher models, and, more specifically, the features in the second layer (P2) of the FPN in both the teacher and student networks exhibit different object perception capabilities. To further bridge this gap, we propose performing semantic alignment at each FPN level between the teacher and the student so that the heterogeneous models generate a consistent feature distribution. More specifically, the features of both networks are firstly standardized to have zero mean and unit variance. Meanwhile, the mean squared error between the standardized features is minimized to better uncover the teacher-student correlation. In addition, this standardization strategy can somewhat reduce the cross-layer difference, allowing both teacher and student networks to comprehensively characterize high-level semantics with consistent representation power. Mathematically, our semantic alignment can be achieved by calculating Pearson Correlation coefficients formulated as follows:


\begin{equation}
P(s,t)=\frac{ {\textstyle \sum_{i=1}^{n}} (s_{i}-\mu  _{t}  )(t_{i}-\mu  _{t})}{\sqrt{{\textstyle \sum_{i=1}^{n}} (s_{i}-\mu  _{s}  )^{2} } \sqrt{{\textstyle \sum_{i=1}^{n}} (t_{i}-\mu  _{t}  )^{2} } }  
\end{equation}
where $P$ is calculated to quantify the degree of correlation between the teacher and student models. $s$ and $t$ represent the teacher and the student feature at each level, respectively, while $\mu$ denotes the mean of a normal distribution. In addition, $n$ denotes the number of FPN levels. Through the feature standardization formulated as above, the teacher and the student features are well aligned to maximize the similarity between pre-standardized features of students and teachers.

\subsection{Loss Function}\label{subsec35}

The overall loss function for training our DFMSD can be formulated as:
\begin{equation}\label{eq6}
L = L_{GT}+ \alpha L_{distill}  
\end{equation}
where $L_{GT}$ is the original detection loss whilst $L_{distill}$ denotes the stage-wise distillation loss as follows:
\begin{equation}\label{eq7}
L_{distill}= \sum_{i=1}^{S-1} \sum_{c=1}^{C} \sum_{h=1}^{H} \sum_{w=1}^{W} (F_{c,h,w}^{T} -\varphi^{S} (F_{Mask}^{S}) ) ^{2}+\beta L_{ME} 
\end{equation}
where $S$ is the number of distillation stages while $C$, $H$, and $W$ represent the channel number, height, width of the feature maps. $F^{S}_{Mask}$ denotes the masked student feature map. In addition, $L_{ME}$ stands for the distillation loss imposed on adaptively augmented data in the last distillation stage:
\begin{equation} \label{equ12}
L_{ME}(\hat{x} ) = \frac{1}{N_{l} } \sum_{c}^{C} \sum_{h}^{H} \sum_{w}^{W} \left \|F_{c,h,w}^{T}(\hat{x} ) -\Phi (F_{Mask}^{S}(\hat{x} ) )  \right \|_{2}^{2}  
\end{equation} 
where $F^{T}_{c,h.w}$($\hat{x}$) and $F^{S}_{Mask}$($\hat{x}$) represent the enhanced masking features generated from the teacher and the student model, respectively. $N_l$ is the total number of elements in the feature map at layer $l$ used for normalization. With the help of Eq.~(\ref{equ12}), our distillation is refined for further improving the performance of the student model. In the above equations, $\alpha$ and $\beta$ are trade-off hyperparameters balancing different terms.




\section{Experiments}\label{sec4}

In this section, we will present comprehensive experiments to evaluate our proposed DFMSD framework after briefly introducing the dataset and experimental setup.

\subsection{Dataset and experimental setup}\label{subsec41}

Our proposed DFMSD method is evaluated in the popular COCO dataset \cite{r23} which comprises over 320k images of 80 different object categories with abundant annotations. It is extensively applied to various tasks, including object detection, image segmentation, and scene understanding. In practice, we use 120k training images for training and 5k validation images for testing. Within our distillation framework, a variety of detectors are involved in our experiments, including RetinaNet~\cite{r8}, FCOS~\cite{r20}, Cascade Mask R-CNN~\cite{r29}, Faster R-CNN~\cite{r32}, GFL~\cite{r54}, RepPoints~\cite{r9}, and Swin-Transformer~\cite{r24}, are involved in our experiments. In particular, we have evaluated our method for heterogeneous distillation in two cases, namely distillation between ViT and CNN architectures and distillation among different CNN detectors. In terms of our SAL strategy, the number of stages are set as $S=2$ for efficiency, which suggests two teacher models are involved for respective distillation stages. For performance measure, we follow~\cite{r90} to adopt Average Precision (AP) and Average Recall (AR) as metrics. All the experiments are conducted on a desktop with an Intel(R) Core(TM) i9-10900K CPU and a 3090 GPU under the PyTorch framework. During the training process, SGD optimizer is used for training all the detectors within 24 epochs. Meanwhile, momentum is set as 0.9 whilst weight decay is set to 0.0001. In addition, a single-scale training strategy is utilized in our experiments. To demonstrate the superiority of our DFMSD model, numerous state-of-the-art (SOTA) masked feature distillation methods are involved in our comparative studies, including FKD~\cite{r53}, FGD~\cite{r15}, MGD~\cite{r42}, AMD~\cite{r43}, DMKD~\cite{r19}, PKD~~\cite{r37} and crossKD~\cite{r38}.

\subsection{Heterogeneous distillation between ViT and CNN Models}\label{subsec42}

In this study, we have conducted extensive experiments in which the advanced Swin-Transformer (ST) model and different categories of CNN detectors are involved. More specifically, ST is used as the teacher framework, while a CNN student model is progressively adapted to the ``weaker'' ST-T model and ``stronger'' ST-S model via our SAL module. All student CNN detectors utilize ResNet50 as the backbone network. According to the CNN detector categories, our experiments for heterogeneous distillation between ViT and CNN models can be categorized into the following three groups.


\subsubsection{Distillation between ST and two-stage CNN detector} 

\begin{table*} 
\centering
\caption{Results of heterogeneous distillation between the ViT and different CNN detectors in the COCO dataset. ST-T and ST-S denote Swin-Transformer-T and Swin-Transformer-S models, respectively. Notably, more powerful ST-S is directly used as the teacher model for MGD and DMKD which are essentially single-stage distillation methods. The best results are highlighted in bold.}
\label{tab1}
\setlength{\tabcolsep}{3.5mm}
\begin{tabular}{c|c|cccc|cccc}
\hline
{\color[HTML]{000000} Teacher}                                                       & {\color[HTML]{000000} Student}                               & {\color[HTML]{000000} mAP}                                     & {\color[HTML]{000000} $AP_s$} & {\color[HTML]{000000} $AP_m$} & {\color[HTML]{000000} $AP_l$} & {\color[HTML]{000000} mAR}                                     & {\color[HTML]{000000} $AR_s$} & {\color[HTML]{000000} $AR_m$} & {\color[HTML]{000000} $AR_l$} \\ \hline
                                                                                     & \begin{tabular}[c]{@{}c@{}}Faster R-CNN (baseline)\end{tabular} & 38.4                                                           & 21.5                       & 42.1                       & 50.3                       & 52.0                                                           & 32.6                       & 55.8                       & 66.1                       \\ 
                                                                                     & MGD                                                          & \begin{tabular}[c]{@{}c@{}}41.9 (+3.5)\end{tabular}          & 23.9                       & 45.8                       & 55.7                       & \begin{tabular}[c]{@{}c@{}}55.5\end{tabular}          & 35.4                       & 60.1                       & 70.7                       \\ 
                                                                                     & DMKD                                                         & \begin{tabular}[c]{@{}c@{}}42.3 (+3.9)\end{tabular}          & 25.2                       & 46.0                       & 55.6                       & \begin{tabular}[c]{@{}c@{}}55.6\end{tabular}          & 35.9                       & 59.3                       & 70.4                       \\ 
                                                                                     & \textbf{DFMSD (ours)}                                       & \textbf{\begin{tabular}[c]{@{}c@{}}43.1 (+4.7)\end{tabular}} & \textbf{25.0}              & \textbf{46.8}              & \textbf{56.8}              & \textbf{\begin{tabular}[c]{@{}c@{}}56.5\end{tabular}} & \textbf{36.6}              & \textbf{60.2}              & \textbf{71.2}              \\ \cline{2-10} 
                                                                                     & \begin{tabular}[c]{@{}c@{}}RetinaNet (baseline)\end{tabular}    & 37.4                                                           & 20.6                       & 40.7                       & 49.7                       & 53.9                                                           & 33.1                       & 57.7                       & 70.2                       \\ 
                                                                                     & MGD                                                          & \begin{tabular}[c]{@{}c@{}}39.7 (+2.3)\end{tabular}          & 22.5                       & 43.4                       & 52.8                       & \begin{tabular}[c]{@{}c@{}}57.0\end{tabular}          & 36.5                       & 61.5                       & 73.3                       \\ 
                                                                                     & DMKD                                                         & \begin{tabular}[c]{@{}c@{}}40.3 (+2.9)\end{tabular}          & 23.3                       & 44.0                       & 53.5                       & \begin{tabular}[c]{@{}c@{}}57.5\end{tabular}          & 38.2                       & 61.8                       & 72.5                       \\ 
                                                                                     & \textbf{DFMSD (ours)}                                       & \textbf{\begin{tabular}[c]{@{}c@{}}41.2 (+3.8)\end{tabular}} & \textbf{24.3}              & \textbf{45.2}              & \textbf{54.4}              & \textbf{\begin{tabular}[c]{@{}c@{}}57.9\end{tabular}} & \textbf{39.4}              & \textbf{62.0}              & \textbf{73.6}              \\ \cline{2-10} 
                                                                                     & \begin{tabular}[c]{@{}c@{}}FCOS (baseline)\end{tabular}        & 35.3                                                           & 20.1                       & 38.3                       & 46.2                       & 53.0                                                           & 32.3                       & 57.4                       & 69.4                       \\ 
                                                                                     & MGD                                                          & \begin{tabular}[c]{@{}c@{}}37.2 (+1.9)\end{tabular}          & 21.2                       & 40.0                       & 49.4                       & \begin{tabular}[c]{@{}c@{}}54.3\end{tabular}          & 33.2                       & 58.8                       & 71.1                       \\ 
                                                                                     & DMKD                                                         & \begin{tabular}[c]{@{}c@{}}37.3 (+2.0)\end{tabular}          & 20.8                       & 40.3                       & 49.3                       & \begin{tabular}[c]{@{}c@{}}54.5\end{tabular}          & 34.2                       & 58.7                       & 70.8                       \\ 
\multirow{-12}{*}{\begin{tabular}[c]{@{}c@{}}ST-T→ST-S\end{tabular}} & \textbf{DFMSD (ours)}                                       & \textbf{\begin{tabular}[c]{@{}c@{}}37.5 (+2.2)\end{tabular}} & \textbf{21.3}              & \textbf{40.4}              & \textbf{49.1}              & \textbf{\begin{tabular}[c]{@{}c@{}}54.7\end{tabular}} & \textbf{34.3}              & \textbf{58.9}              & \textbf{71.1}              \\ \hline
\end{tabular}
\end{table*}

In this group of experiments, the Faster R-CNN detector with ResNet50 backbone serves as the student model. As demonstrated in Table~\ref{tab1}, our DFMSD method significantly improves baseline by 4.7\% mAP, reporting the highest precision at 43.1\%. Moreover, it surpasses the SOTA methods MGD and DMKD by 1.2\% and 0.8\%, respectively. Similar performance improvements are also observed in the mAR metric. These results fully demonstrate that our method can take advantage of stage-wise distillation to achieve more performance gains for the student model compared to the single-stage distillation approaches like MGD and DMKD.

\subsubsection{Distillation between ST and one-stage CNN detector}
Different from the first-group experimental setup, the student Faster R-CNN framework is replaced by the RetinaNet framework. Similar to the results of the first group, our method provides a significant improvement over baseline in mAP performance by 3.8\% and mAR performance by 4.0\%. Furthermore, the proposed DFMSD consistently outperforms the other two competitors and particularly beats its predecessor DMKD by 0.9\% mAP, which suggests considerable advantages of our model.

\subsubsection{Distillation between ST and anchor-free CNN detector}
To further assess the generalizability of our proposed method, the anchor-free FCOS detector is used as the student network. Although our model reports less performance improvements compared with the previous two groups, it still exhibits consistent performance advantages.


\begin{table*}[h] 
\centering
\caption{Results of heterogeneous distillation in the COCO dataset when the two-stage Cascade Mask R-CNN is used as the teacher framework and other types of CNN detectors are employed for the student models. Notably, Cascade Mask R-CNN with a ``stronger'' backbone ResNeXt101 is directly used as the teacher model for MGD and DMKD which are essentially single-stage distillation methods. The best results are highlighted in bold.}
\label{tab2}
\setlength{\tabcolsep}{3.5mm}
\begin{tabular}{c|c|cccc|cccc}
\hline
{\color[HTML]{000000} Teacher}                                                                                                & {\color[HTML]{000000} Student}                                                    & {\color[HTML]{000000} mAP}                                                            & {\color[HTML]{000000} $AP_s$}           & {\color[HTML]{000000} $AP_m$}           & {\color[HTML]{000000} $AP_l$}           & {\color[HTML]{000000} mAR}                                                            & {\color[HTML]{000000} $AR_s$}           & {\color[HTML]{000000} $AR_m$}           & {\color[HTML]{000000} $AR_l$}           \\ \hline
{\color[HTML]{000000} }                                                                                                       & {\color[HTML]{000000} \begin{tabular}[c]{@{}c@{}}RetinaNet (baseline)\end{tabular}} & {\color[HTML]{000000} 37.4}                                                           & {\color[HTML]{000000} 20.6}          & {\color[HTML]{000000} 40.7}          & {\color[HTML]{000000} 49.7}          & {\color[HTML]{000000} 53.9}                                                           & {\color[HTML]{000000} 33.1}          & {\color[HTML]{000000} 57.7}          & {\color[HTML]{000000} 70.2}          \\
{\color[HTML]{000000} }                                                                                                       & {\color[HTML]{000000} MGD}                                                        & {\color[HTML]{000000} \begin{tabular}[c]{@{}c@{}}39.1 (+1.7)\end{tabular}}          & {\color[HTML]{000000} 21.6}          & {\color[HTML]{000000} 42.7}          & {\color[HTML]{000000} 52.2}          & {\color[HTML]{000000} \begin{tabular}[c]{@{}c@{}}56.4\end{tabular}}          & {\color[HTML]{000000} 36.7}          & {\color[HTML]{000000} 60.6}          & {\color[HTML]{000000} 72.6}          \\
{\color[HTML]{000000} }                                                                                                       & {\color[HTML]{000000} DMKD}                                                       & {\color[HTML]{000000} \begin{tabular}[c]{@{}c@{}}39.7 (+2.3)\end{tabular}}          & {\color[HTML]{000000} 21.2}          & {\color[HTML]{000000} 43.5}          & {\color[HTML]{000000} 52.8}          & {\color[HTML]{000000} \begin{tabular}[c]{@{}c@{}}56.5\end{tabular}}          & {\color[HTML]{000000} 36.6}          & {\color[HTML]{000000} 60.6}          & {\color[HTML]{000000} 73.3}          \\
{\color[HTML]{000000} }                                                                                                       & {\color[HTML]{000000} \textbf{DFMSD(ours)}}                                     & {\color[HTML]{000000} \textbf{\begin{tabular}[c]{@{}c@{}}40.1 (+2.7)\end{tabular}}} & {\color[HTML]{000000} \textbf{22.1}} & {\color[HTML]{000000} \textbf{43.4}} & {\color[HTML]{000000} \textbf{54.0}} & {\color[HTML]{000000} \textbf{\begin{tabular}[c]{@{}c@{}}56.9\end{tabular}}} & {\color[HTML]{000000} \textbf{35.5}} & {\color[HTML]{000000} \textbf{60.6}} & {\color[HTML]{000000} \textbf{73.5}} \\ \cline{2-10}
{\color[HTML]{000000} }                                                                                                       & {\color[HTML]{000000} \begin{tabular}[c]{@{}c@{}}FCOS (baseline)\end{tabular}} & {\color[HTML]{000000} 35.3}                                                           & {\color[HTML]{000000} 20.1}          & {\color[HTML]{000000} 38.3}          & {\color[HTML]{000000} 46.2}          & {\color[HTML]{000000} 53.0}                                                           & {\color[HTML]{000000} 32.3}          & {\color[HTML]{000000} 57.4}          & {\color[HTML]{000000} 69.4}          \\
{\color[HTML]{000000} }                                                                                                       & {\color[HTML]{000000} MGD}                                                        & {\color[HTML]{000000} \begin{tabular}[c]{@{}c@{}}35.4 (+0.1)\end{tabular}}          & {\color[HTML]{000000} 20.8}          & {\color[HTML]{000000} 38.7}          & {\color[HTML]{000000} 45.7}          & {\color[HTML]{000000} \begin{tabular}[c]{@{}c@{}}53.3\end{tabular}}          & {\color[HTML]{000000} 32.7}          & {\color[HTML]{000000} 58.0}          & {\color[HTML]{000000} 69.0}          \\
{\color[HTML]{000000} }                                                                                                       & {\color[HTML]{000000} DMKD}                                                       & {\color[HTML]{000000} \begin{tabular}[c]{@{}c@{}}36.2 (+0.9)\end{tabular}}          & {\color[HTML]{000000} 20.5}          & {\color[HTML]{000000} 39.5}          & {\color[HTML]{000000} 47.2}          & {\color[HTML]{000000} \begin{tabular}[c]{@{}c@{}}53.8\end{tabular}}          & {\color[HTML]{000000} 32.9}          & {\color[HTML]{000000} 58.0}          & {\color[HTML]{000000} 70.3}          \\
\multirow{-8}{*}{{\color[HTML]{000000} \begin{tabular}[c]{@{}c@{}}Cascade\\ Mask R-CNN\\ ResNet101→\\ ResNeXt101\end{tabular}}} & {\color[HTML]{000000} \textbf{DFMSD(ours)}}                                     & {\color[HTML]{000000} \textbf{\begin{tabular}[c]{@{}c@{}}36.8 (+1.5)\end{tabular}}} & {\color[HTML]{000000} \textbf{21.1}} & {\color[HTML]{000000} \textbf{39.7}} & {\color[HTML]{000000} \textbf{47.6}} & {\color[HTML]{000000} \textbf{\begin{tabular}[c]{@{}c@{}}54.5\end{tabular}}} & {\color[HTML]{000000} \textbf{34.1}} & {\color[HTML]{000000} \textbf{58.7}} & {\color[HTML]{000000} \textbf{70.9}} \\ \hline
\end{tabular}
\end{table*}

\begin{table*}[h] 
\centering
\caption{Results of heterogeneous distillation in the COCO dataset when one-stage CNN models are used as the teacher detectors and other types of CNN detectors are employed for the student models. The best results are highlighted in bold.}
\label{tab3}
\setlength{\tabcolsep}{3.5mm}
\begin{tabular}{c|c|cccc|cccc}
\hline
{\color[HTML]{000000} Teacher}                                                                             & {\color[HTML]{000000} Student}                                                      & {\color[HTML]{000000} mAP}                                                            & {\color[HTML]{000000} $AP_s$}           & {\color[HTML]{000000} $AP_m$}           & {\color[HTML]{000000} $AP_l$}           & {\color[HTML]{000000} mAR}                                                            & {\color[HTML]{000000} $AR_s$}           & {\color[HTML]{000000} $AR_m$}           & {\color[HTML]{000000} $AR_l$}           \\ \hline
{\color[HTML]{000000} }                                                                                    & {\color[HTML]{000000} \begin{tabular}[c]{@{}l@{}}Faster R-CNN (baseline)\end{tabular}} & {\color[HTML]{000000} 38.4}                                                           & {\color[HTML]{000000} 21.5}          & {\color[HTML]{000000} 42.1}          & {\color[HTML]{000000} 50.3}          & {\color[HTML]{000000} 52.0}                                                           & {\color[HTML]{000000} 32.6}          & {\color[HTML]{000000} 55.8}          & {\color[HTML]{000000} 66.1}          \\
{\color[HTML]{000000} }                                                                                    & {\color[HTML]{000000} MGD}                                                          & {\color[HTML]{000000} \begin{tabular}[c]{@{}l@{}}40.9 (+2.5)\end{tabular}}          & {\color[HTML]{000000} 23.5}          & {\color[HTML]{000000} 44.8}          & {\color[HTML]{000000} 541}           & {\color[HTML]{000000} \begin{tabular}[c]{@{}l@{}}54.0\end{tabular}}          & {\color[HTML]{000000} 33.9}          & {\color[HTML]{000000} 58.2}          & {\color[HTML]{000000} 69.1}          \\
{\color[HTML]{000000} }                                                                                    & {\color[HTML]{000000} DMKD}                                                         & {\color[HTML]{000000} \begin{tabular}[c]{@{}l@{}}40.8 (+2.4)\end{tabular}}          & {\color[HTML]{000000} 22.9}          & {\color[HTML]{000000} 44.7}          & {\color[HTML]{000000} 54.2}          & {\color[HTML]{000000} \begin{tabular}[c]{@{}l@{}}53.9\end{tabular}}          & {\color[HTML]{000000} 33.6}          & {\color[HTML]{000000} 57.6}          & {\color[HTML]{000000} 68.9}          \\
{\color[HTML]{000000} }                                                                                    & {\color[HTML]{000000} \textbf{DFMSD(ours)}}                                       & {\color[HTML]{000000} \textbf{\begin{tabular}[c]{@{}l@{}}41.2 (+2.8)\end{tabular}}} & {\color[HTML]{000000} \textbf{23.5}} & {\color[HTML]{000000} \textbf{45.2}} & {\color[HTML]{000000} \textbf{54.4}} & {\color[HTML]{000000} \textbf{\begin{tabular}[c]{@{}l@{}}54.1\end{tabular}}} & {\color[HTML]{000000} \textbf{34.2}} & {\color[HTML]{000000} \textbf{58.2}} & {\color[HTML]{000000} \textbf{68.7}} \\  \cline{2-10}
{\color[HTML]{000000} }                                                                                    & {\color[HTML]{000000} \begin{tabular}[c]{@{}l@{}}FCOS (baseline)\end{tabular}}        & {\color[HTML]{000000} 35.3}                                                           & {\color[HTML]{000000} 20.1}          & {\color[HTML]{000000} 38.3}          & {\color[HTML]{000000} 46.2}          & {\color[HTML]{000000} 53.0}                                                           & {\color[HTML]{000000} 32.3}          & {\color[HTML]{000000} 57.4}          & {\color[HTML]{000000} 69.4}          \\
{\color[HTML]{000000} }                                                                                    & {\color[HTML]{000000} MGD}                                                          & {\color[HTML]{000000} \begin{tabular}[c]{@{}l@{}}35.4 (+0.1)\end{tabular}}          & {\color[HTML]{000000} 20.7}          & {\color[HTML]{000000} 38.7}          & {\color[HTML]{000000} 45.6}          & {\color[HTML]{000000} \begin{tabular}[c]{@{}l@{}}53.2\end{tabular}}          & {\color[HTML]{000000} 32.8}          & {\color[HTML]{000000} 58.0}          & {\color[HTML]{000000} 68.9}          \\
{\color[HTML]{000000} }                                                                                    & {\color[HTML]{000000} DMKD}                                                         & {\color[HTML]{000000} \begin{tabular}[c]{@{}l@{}}36.5 (+1.2)\end{tabular}}          & {\color[HTML]{000000} 20.8}          & {\color[HTML]{000000} 39.5}          & {\color[HTML]{000000} 47.6}          & {\color[HTML]{000000} \begin{tabular}[c]{@{}l@{}}53.9\end{tabular}}          & {\color[HTML]{000000} 33.1}          & {\color[HTML]{000000} 58.3}          & {\color[HTML]{000000} 70.0}          \\
\multirow{-8}{*}{{\color[HTML]{000000} \begin{tabular}[c]{@{}c@{}}RetinaNet\\ResNet101→\\ResNeXt101\end{tabular}}} & {\color[HTML]{000000} \textbf{DFMSD(ours)}}                                       & {\color[HTML]{000000} \textbf{\begin{tabular}[c]{@{}l@{}}36.8 (+1.5)\end{tabular}}} & {\color[HTML]{000000} \textbf{21.0}} & {\color[HTML]{000000} \textbf{39.5}} & {\color[HTML]{000000} \textbf{48.1}} & {\color[HTML]{000000} \textbf{\begin{tabular}[c]{@{}l@{}}54.1\end{tabular}}} & {\color[HTML]{000000} \textbf{34.1}} & {\color[HTML]{000000} \textbf{58.3}} & {\color[HTML]{000000} \textbf{70.1}} \\  \hline
{\color[HTML]{000000} }                                                                                    & {\color[HTML]{000000} \begin{tabular}[c]{@{}l@{}}FCOS (baseline)\end{tabular}}        & {\color[HTML]{000000} 35.3}                                                           & {\color[HTML]{000000} 20.1}          & {\color[HTML]{000000} 38.3}          & {\color[HTML]{000000} 46.2}          & {\color[HTML]{000000} 53.0}                                                           & {\color[HTML]{000000} 32.3}          & {\color[HTML]{000000} 57.4}          & {\color[HTML]{000000} 69.4}          \\
{\color[HTML]{000000} }                                                                                    & {\color[HTML]{000000} MGD}                                                          & {\color[HTML]{000000} \begin{tabular}[c]{@{}l@{}}35.9 (+0.6)\end{tabular}}          & {\color[HTML]{000000} 20.4}          & {\color[HTML]{000000} 38.9}          & {\color[HTML]{000000} 46.6}          & {\color[HTML]{000000} \begin{tabular}[c]{@{}l@{}}53.4\end{tabular}}          & {\color[HTML]{000000} 32.8}          & {\color[HTML]{000000} 57.7}          & {\color[HTML]{000000} 69.5}          \\
{\color[HTML]{000000} }                                                                                    & {\color[HTML]{000000} DMKD}                                                         & {\color[HTML]{000000} \begin{tabular}[c]{@{}l@{}}36.7 (+1.4)\end{tabular}}          & {\color[HTML]{000000} 20.9}          & {\color[HTML]{000000} 39.9}          & {\color[HTML]{000000} 47.3}          & {\color[HTML]{000000} \begin{tabular}[c]{@{}l@{}}53.9\end{tabular}}          & {\color[HTML]{000000} 33.3}          & {\color[HTML]{000000} 58.3}          & {\color[HTML]{000000} 69.9}          \\
\multirow{-4}{*}{{\color[HTML]{000000} \begin{tabular}[c]{@{}c@{}}GFL\\ResNet101→\\ResNeXt101\end{tabular}}}       & {\color[HTML]{000000} \textbf{DFMSD(ours)}}                                       & {\color[HTML]{000000} \textbf{\begin{tabular}[c]{@{}l@{}}36.9 (+1.6)\end{tabular}}} & {\color[HTML]{000000} \textbf{21.2}} & {\color[HTML]{000000} \textbf{39.9}} & {\color[HTML]{000000} \textbf{48.1}} & {\color[HTML]{000000} \textbf{\begin{tabular}[c]{@{}l@{}}54.2\end{tabular}}} & {\color[HTML]{000000} \textbf{33.8}} & {\color[HTML]{000000} \textbf{58.7}} & {\color[HTML]{000000} \textbf{70.3}} \\ \hline
\end{tabular}
\end{table*}

\begin{table*}[h] 
\centering
\caption{Results of heterogeneous distillation in the COCO dataset when anchor-free FCOS is used as the teacher model and other types of CNN detectors are employed for the student models. The best results are highlighted in bold.}
\label{tab4}
\setlength{\tabcolsep}{3.5mm}
\begin{tabular}{c|c|cccc|cccc}
\hline
{\color[HTML]{000000} Teacher}                                                                         & {\color[HTML]{000000} Student}                                                      & {\color[HTML]{000000} mAP}                                                            & {\color[HTML]{000000} $AP_s$}           & {\color[HTML]{000000} $AP_m$}           & {\color[HTML]{000000} $AP_l$}           & {\color[HTML]{000000} mAR}                                                            & {\color[HTML]{000000} $AR_s$}           & {\color[HTML]{000000} $AR_m$}           & {\color[HTML]{000000} $AR_l$}           \\ \hline
{\color[HTML]{000000} }                                                                                & {\color[HTML]{000000} \begin{tabular}[c]{@{}l@{}}Faster R-CNN (baseline)\end{tabular}} & {\color[HTML]{000000} 38.4}                                                           & {\color[HTML]{000000} 21.5}          & {\color[HTML]{000000} 42.1}          & {\color[HTML]{000000} 50.3}          & {\color[HTML]{000000} 52.0}                                                           & {\color[HTML]{000000} 32.6}          & {\color[HTML]{000000} 55.8}          & {\color[HTML]{000000} 66.1}          \\
{\color[HTML]{000000} }                                                                                & {\color[HTML]{000000} MGD}                                                          & {\color[HTML]{000000} \begin{tabular}[c]{@{}l@{}}40.1 (+1.7)\end{tabular}}          & {\color[HTML]{000000} 23.1}          & {\color[HTML]{000000} 44.1}          & {\color[HTML]{000000} 52.2}          & {\color[HTML]{000000} \begin{tabular}[c]{@{}l@{}}53.6\end{tabular}}          & {\color[HTML]{000000} 34.5}          & {\color[HTML]{000000} 57.3}          & {\color[HTML]{000000} 67.9}          \\
{\color[HTML]{000000} }                                                                                & {\color[HTML]{000000} DMKD}                                                         & {\color[HTML]{000000} \begin{tabular}[c]{@{}l@{}}40.5 (+2.1)\end{tabular}}          & {\color[HTML]{000000} 22.7}          & {\color[HTML]{000000} 44.6}          & {\color[HTML]{000000} 52.7}          & {\color[HTML]{000000} \begin{tabular}[c]{@{}l@{}}53.7\end{tabular}}          & {\color[HTML]{000000} 33.8}          & {\color[HTML]{000000} 57.9}          & {\color[HTML]{000000} 67.7}          \\
{\color[HTML]{000000} }                                                                                & {\color[HTML]{000000} \textbf{DFMSD(ours)}}                                       & {\color[HTML]{000000} \textbf{\begin{tabular}[c]{@{}l@{}}40.8 (+2.4)\end{tabular}}} & {\color[HTML]{000000} \textbf{22.9}} & {\color[HTML]{000000} \textbf{44.5}} & {\color[HTML]{000000} \textbf{54.0}} & {\color[HTML]{000000} \textbf{\begin{tabular}[c]{@{}l@{}}53.9\end{tabular}}} & {\color[HTML]{000000} \textbf{33.6}} & {\color[HTML]{000000} \textbf{57.5}} & {\color[HTML]{000000} \textbf{68.9}} \\  \cline{2-10}
{\color[HTML]{000000} }                                                                                & {\color[HTML]{000000} \begin{tabular}[c]{@{}l@{}}GFL (baseline)\end{tabular}}         & {\color[HTML]{000000} 39.6}                                                           & {\color[HTML]{000000} 22.3}          & {\color[HTML]{000000} 43.6}          & {\color[HTML]{000000} 52.2}          & {\color[HTML]{000000} 58.2}                                                           & {\color[HTML]{000000} 35.8}          & {\color[HTML]{000000} 63.1}          & {\color[HTML]{000000} 75.2}          \\
{\color[HTML]{000000} }                                                                                & {\color[HTML]{000000} MGD}                                                          & {\color[HTML]{000000} \begin{tabular}[c]{@{}l@{}}40.1 (+0.5)\end{tabular}}          & {\color[HTML]{000000} 22.8}          & {\color[HTML]{000000} 43.8}          & {\color[HTML]{000000} 52.9}          & {\color[HTML]{000000} \begin{tabular}[c]{@{}l@{}}58.8\end{tabular}}          & {\color[HTML]{000000} 36.8}          & {\color[HTML]{000000} 63.5}          & {\color[HTML]{000000} 75.8}          \\
{\color[HTML]{000000} }                                                                                & {\color[HTML]{000000} DMKD}                                                         & {\color[HTML]{000000} \begin{tabular}[c]{@{}l@{}}40.4 (+0.8)\end{tabular}}          & {\color[HTML]{000000} 22.4}          & {\color[HTML]{000000} 44.3}          & {\color[HTML]{000000} 53.8}          & {\color[HTML]{000000} \begin{tabular}[c]{@{}l@{}}59.1\end{tabular}}          & {\color[HTML]{000000} 37.8}          & {\color[HTML]{000000} 63.8}          & {\color[HTML]{000000} 75.6}          \\
{\color[HTML]{000000} }                                                                                & {\color[HTML]{000000} \textbf{DFMSD(ours)}}                                       & {\color[HTML]{000000} \textbf{\begin{tabular}[c]{@{}l@{}}40.7 (+1.1)\end{tabular}}} & {\color[HTML]{000000} \textbf{22.8}} & {\color[HTML]{000000} \textbf{44.5}} & {\color[HTML]{000000} \textbf{53.1}} & {\color[HTML]{000000} \textbf{\begin{tabular}[c]{@{}l@{}}59.6\end{tabular}}} & {\color[HTML]{000000} \textbf{38.3}} & {\color[HTML]{000000} \textbf{64.6}} & {\color[HTML]{000000} \textbf{76.1}} \\  \cline{2-10}
{\color[HTML]{000000} }                                                                                & {\color[HTML]{000000} \begin{tabular}[c]{@{}l@{}}RetinaNet (baseline)\end{tabular}}   & {\color[HTML]{000000} 37.4}                                                           & {\color[HTML]{000000} 20.6}          & {\color[HTML]{000000} 40.7}          & {\color[HTML]{000000} 49.7}          & {\color[HTML]{000000} 53.9}                                                           & {\color[HTML]{000000} 33.1}          & {\color[HTML]{000000} 57.7}          & {\color[HTML]{000000} 70.2}          \\
{\color[HTML]{000000} }                                                                                & {\color[HTML]{000000} MGD}                                                          & {\color[HTML]{000000} \begin{tabular}[c]{@{}l@{}}39.5 (+2.1)\end{tabular}}          & {\color[HTML]{000000} 22.1}          & {\color[HTML]{000000} 43.0}          & {\color[HTML]{000000} 52.2}          & {\color[HTML]{000000} \begin{tabular}[c]{@{}l@{}}56.5\end{tabular}}          & {\color[HTML]{000000} 37.6}          & {\color[HTML]{000000} 60.4}          & {\color[HTML]{000000} 72.0}          \\
{\color[HTML]{000000} }                                                                                & {\color[HTML]{000000} DMKD}                                                         & {\color[HTML]{000000} \begin{tabular}[c]{@{}l@{}}39.7 (+2.3)\end{tabular}}          & {\color[HTML]{000000} 22.4}          & {\color[HTML]{000000} 42.9}          & {\color[HTML]{000000} 52.2}          & {\color[HTML]{000000} \begin{tabular}[c]{@{}l@{}}56.9\end{tabular}}          & {\color[HTML]{000000} 36.7}          & {\color[HTML]{000000} 60.8}          & {\color[HTML]{000000} 72.6}          \\
\multirow{-12}{*}{{\color[HTML]{000000} \begin{tabular}[c]{@{}c@{}}FCOS \\ ResNet101→\\ ResNeXt101\end{tabular}}} & {\color[HTML]{000000} \textbf{DFMSD(ours)}}                                       & {\color[HTML]{000000} \textbf{\begin{tabular}[c]{@{}l@{}}40.2 (+2.8)\end{tabular}}} & {\color[HTML]{000000} \textbf{22.5}} & {\color[HTML]{000000} \textbf{44.0}} & {\color[HTML]{000000} \textbf{53.3}} & {\color[HTML]{000000} \textbf{\begin{tabular}[c]{@{}l@{}}56.9\end{tabular}}} & {\color[HTML]{000000} \textbf{37.4}} & {\color[HTML]{000000} \textbf{60.8}} & {\color[HTML]{000000} \textbf{72.6}} \\ \hline
\end{tabular}
\end{table*}

\subsection{Heterogeneous distillation among CNN models}\label{subsec43}

In addition to the distillation between the ViT and CNN architectures, we have carried out additional heterogeneous distillation experiments among different categories of CNN detectors, namely two-stage models, one-stage models, and anchor-free models. The experiments are presented as the following three groups. Consistent with the above experiments, all the student CNN detectors adopt ResNet-50 as the backbone network.

\subsubsection{Distillation using two-stage detectors as the teachers}

In this group of experiment, two-stage Cascade Mask R-CNN is used for the teacher framework while the other CNN models for the students. In particular, the ``weaker'' and the ``stronger'' teacher models are Cascade Mask R-CNN with backbone networks used as respective ResNet-50 and ResNext-101. As revealed in Table~\ref{tab2}, our distillation method significantly improves the one-stage student detector RetinaNet by 2.7\%, reporting highest 40.1\% mAP. Meanwhile, our method outperforms MGD and DMKD by 1\% and 0.4\% mAP respectively, which demonstrates our distillation scheme is more helpful for improving the student model. When the anchor-free FCOS detector is used for the student model while the Cascade Mask R-CNN remains the teacher network, the proposed DFMSD improves the baseline by 1.5\% mAP with fewer performance gains compared to the above experiments, whereas the best results are still achieved by our method.

\subsubsection{Distillation using one-stage detectors as the teachers}

When using a one-stage detector as the teacher model, the RetinaNet frameworks with a ``weaker'' backbone ResNet-101 and a ``stronger'' backbone ResNeXt-101 are firstly used for the successive distillation stages. As demonstrated in Table.~\ref{tab3}, our proposed DFMSD achieves respective performance boosts of 2.8\% and 1.5\% over the baseline student models of Faster R-CNN and FCOS, and outperforms the other two single-stage distillation approaches with consistent performance advantages in mAP and mAR. When the teacher framework is replaced by a more powerful GFL detector~\cite{r73} while the student network is used as the FCOS~\cite{r20}, similar improvements can also be observed over both the baseline and the other competitors, which implies that the student models can benefit from our distillation scheme with effective knowledge transfer.

\subsubsection{Distillation using anchor-free detectors as the teachers}
When adopting the anchor-free detector as the teacher network, FCOS is used as the teacher model, while three different types of detectors are used as the student models, namely two-stage Faster R-CNN, as well as one-stage GFL and RetinaNet. With Faster R-CNN as the student model, it is shown in Table~\ref{tab4} that the performance boosts over the baseline achieved by our method reach 2.4\% mAP and 1.9\% mAR, which consistently beats the other distillation approaches. When one-stage student detectors are involved, including GFL and RetinaNet, our distillation method still achieves the best results. In particular, the proposed DFMSD elevates the mAP accuracy of RetinaNet from 37. 4\% to 40. 2\% and the mAR accuracies from 53. 9\% to 56. 9\%, demonstrating significant performance improvements. In addition, our DFMSD is also superior to MGD and DMKD with consistent improvements exceeding 0.5\%. The results unanimously showcase the framework-independent advantages of our method in various cases, suggesting that more crucial information can be learned from diverse heterogeneous teacher models with the help of our distillation paradigm for improving the student performance.





\subsection{Comparison with SOTA Heterogeneous Knowledge Distillation Methods}\label{subsec44}

To further demonstrate the superiority of our method, we compare the proposed DFMSD with the other heterogeneous distillation approaches including PKD and crossKD. In particular, crossKD adopts a similar adaptive cross-head approach which aims at facilitating the prediction imitation to bridge the gap between teachers and students. In practice, our DFMSD performs stage-wise distillation such that the RetinaNet student detector with ResNet50 backbone network can adaptively learn from the original ``weaker'' Swin-Transformer-T (ST-T) to the ``stronger'' Swin-Transformer-S (ST-S). In contrast, PKD and crossKD, which are single-distillation methods without feature masking, function by directly transferring knowledge from Cascade Mask R-CNN to ST-T. As revealed in Table~\ref{tab5}, our method outperforms both PKD and crossKD by respective 1.3\% and 0.6\% mAP accuracies, which indicates that a simple cross-head strategy is insufficient to reduce the difference between heterogeneous teacher and student models, and thus exhibits suboptimal performance.


\begin{table*}[h] 
\centering
\caption{Comparison of our method and the SOTA heterogeneous knowledge distillation methods in COCO dataset. The best results are highlighted in bold.}
\label{tab5}
\setlength{\tabcolsep}{3.5mm}
\begin{tabular}{c|c|cccc}
\hline
{\color[HTML]{000000} Teacher}                                                                            & {\color[HTML]{000000} Student}                                                    & {\color[HTML]{000000} mAP}                                                            & {\color[HTML]{000000} $AP_s$}           & {\color[HTML]{000000} $AP_m$}           & {\color[HTML]{000000} $AP_l$}           \\ \hline
{\color[HTML]{000000} }                                                                                   & {\color[HTML]{000000} \begin{tabular}[c]{@{}c@{}}RetinaNet (baseline)\end{tabular}} & {\color[HTML]{000000} 37.4}                                                           & {\color[HTML]{000000} 20.6}          & {\color[HTML]{000000} 40.7}          & {\color[HTML]{000000} 49.7}          \\
{\color[HTML]{000000} }                                                                                   & {\color[HTML]{000000} PKD}                                                        & {\color[HTML]{000000} \begin{tabular}[c]{@{}c@{}}38.6 (+1.2)\end{tabular}}          & {\color[HTML]{000000} 22.2}          & {\color[HTML]{000000} 42.1}          & {\color[HTML]{000000} 49.9}          \\
{\color[HTML]{000000} }                                                                                   & {\color[HTML]{000000} DMKD}                                                       & {\color[HTML]{000000} \begin{tabular}[c]{@{}c@{}}38.7 (+1.3)\end{tabular}}          & {\color[HTML]{000000} 22.5}          & {\color[HTML]{000000} 42.3}          & {\color[HTML]{000000} 50.6}          \\
{\color[HTML]{000000} }                                                                                   & {\color[HTML]{000000} CrossKD}                                                    & {\color[HTML]{000000} \begin{tabular}[c]{@{}c@{}}39.4 (+2.0)\end{tabular}}          & {\color[HTML]{000000} 22.6}          & {\color[HTML]{000000} 43.3}          & {\color[HTML]{000000} 51.3}          \\
\multirow{-5}{*}{{\color[HTML]{000000} \begin{tabular}[c]{@{}c@{}}Cascade
Mask R-CNN (ResNet101)\\→ST-T\end{tabular}}} & {\color[HTML]{000000} \textbf{DFMSD(ours)}}                                     & {\color[HTML]{000000} \textbf{\begin{tabular}[c]{@{}c@{}}40.0 (+2.6)\end{tabular}}} & {\color[HTML]{000000} \textbf{22.7}} & {\color[HTML]{000000} \textbf{43.9}} & {\color[HTML]{000000} \textbf{52.4}} \\ \hline
\end{tabular}
\end{table*}


\subsection{Experiments of homogeneous distillation}\label{subsec45}

In addition to the aforementioned heterogeneous distillation experiments, we have also evaluated our method in the case of homogeneous distillation for detection and compared it with the other SOTA schemes in the COCO, including FKD, FGD, MGD, AMD, and DMKD. In homogeneous distillation, the teacher and student models share the same detection framework, whereas the former has a more powerful backbone network than the latter. As shown in Table~\ref{tab6}, four different detectors, including RetinaNet, RepPoints, GFL and FCOS are involved in our comparative studies. In addition, the backbone networks of the teacher and student frameworks are used as ResNeXt101 and ResNet50, respectively. The only exception is our DFMSD framework which incorporates two teacher models with respective ResNet101 and ResNeXt101 backbones in the process of stage-wise adaptive learning. The pre-trained models for the teacher are directly borrowed from the MMDetection toolbox~\cite{r51}. It can be observed from the results that our DFMSD achieves consistent superiority to all the competing methods. For example, when the RetinaNet is used as the detection framework, our approach outperforms its predecessor DMKD by 0.5\% mAP and beats the other signle-distillation methods. When using a more advanced GFL detector, the performance advantage against DMKD reaches 1.4\%, which demonstrates the substantial benefits of integrating the stage-wise distillation mechanism into the feature masking framework. 



\begin{table*}[h] 
\centering
\caption{Comparison of different methods for homogeneous knowledge distillation in the COCO dataset. Different from the other distillation methods in which a single teacher with ResNeXt101 backbone is used, our approach leverages dual teachers with both ``weaker'' ResNet101 and ``stronger'' ResNeXt101 backbones for our SAL module.}
\label{tab6}
\setlength{\tabcolsep}{3.5mm}
\begin{tabular}{c|c|cccc|cccc}
\hline
{\color[HTML]{000000} \textbf{Teacher}}                                                                           & {\color[HTML]{000000} \textbf{Student}}                                             & {\color[HTML]{000000} \textbf{mAP}}                                                   & {\color[HTML]{000000} \textbf{$AP_s$}}  & {\color[HTML]{000000} \textbf{$AP_m$}}  & {\color[HTML]{000000} \textbf{$AP_l$}}  & {\color[HTML]{000000} \textbf{mAR}}                                                   & {\color[HTML]{000000} \textbf{$AR_s$}}  & {\color[HTML]{000000} \textbf{$AR_m$}}  & {\color[HTML]{000000} \textbf{$AR_l$}}  \\ \hline
{\color[HTML]{000000} }                                                                                           & {\color[HTML]{000000} \begin{tabular}[c]{@{}c@{}}RetinaNet (baseline)\end{tabular}}   & {\color[HTML]{000000} 37.4}                                                           & {\color[HTML]{000000} 20.6}          & {\color[HTML]{000000} 40.7}          & {\color[HTML]{000000} 49.7}          & {\color[HTML]{000000} 53.9}                                                           & {\color[HTML]{000000} 33.1}          & {\color[HTML]{000000} 57.7}          & {\color[HTML]{000000} 70.2}          \\
{\color[HTML]{000000} }                                                                                           & {\color[HTML]{000000} FKD}                                                          & {\color[HTML]{000000} \begin{tabular}[c]{@{}c@{}}39.6 (+2.2)\end{tabular}}          & {\color[HTML]{000000} 22.7}          & {\color[HTML]{000000} 43.4}          & {\color[HTML]{000000} 52.5}          & {\color[HTML]{000000} \begin{tabular}[c]{@{}c@{}}56.1\end{tabular}}          & {\color[HTML]{000000} 36.8}          & {\color[HTML]{000000} 60.0}          & {\color[HTML]{000000} 72.1}          \\
{\color[HTML]{000000} }                                                                                           & {\color[HTML]{000000} FGD}                                                          & {\color[HTML]{000000} \begin{tabular}[c]{@{}c@{}}40.7 (+3.3)\end{tabular}}          & {\color[HTML]{000000} 22.9}          & {\color[HTML]{000000} 45.0}          & {\color[HTML]{000000} 54.7}          & {\color[HTML]{000000} \begin{tabular}[c]{@{}c@{}}56.8\end{tabular}}          & {\color[HTML]{000000} 36.5}          & {\color[HTML]{000000} 61.4}          & {\color[HTML]{000000} 72.8}          \\
{\color[HTML]{000000} }                                                                                           & {\color[HTML]{000000} MGD}                                                          & {\color[HTML]{000000} \begin{tabular}[c]{@{}c@{}}41.0 (+3.6)\end{tabular}}          & {\color[HTML]{000000} 23.4}          & {\color[HTML]{000000} 45.3}          & {\color[HTML]{000000} 55.7}          & {\color[HTML]{000000} \begin{tabular}[c]{@{}c@{}}57.0\end{tabular}}          & {\color[HTML]{000000} 37.2}          & {\color[HTML]{000000} 61.7}          & {\color[HTML]{000000} 72.8}          \\
{\color[HTML]{000000} }                                                                                           & {\color[HTML]{000000} AMD}                                                          & {\color[HTML]{000000} \begin{tabular}[c]{@{}c@{}}41.3 (+3.9)\end{tabular}}          & {\color[HTML]{000000} 23.9}          & {\color[HTML]{000000} 45.4}          & {\color[HTML]{000000} 55.7}          & {\color[HTML]{000000} \begin{tabular}[c]{@{}c@{}}57.4\end{tabular}}          & {\color[HTML]{000000} 38.2}          & {\color[HTML]{000000} 61.7}          & {\color[HTML]{000000} 73.5}          \\
{\color[HTML]{000000} }                                                                                           & {\color[HTML]{000000} DMKD}                                                         & {\color[HTML]{000000} \begin{tabular}[c]{@{}c@{}}41.5 (+4.1)\end{tabular}}          & {\color[HTML]{000000} 24.0}          & {\color[HTML]{000000} 45.8}          & {\color[HTML]{000000} 55.8}          & {\color[HTML]{000000} \begin{tabular}[c]{@{}c@{}}57.8\end{tabular}}          & {\color[HTML]{000000} 38.5}          & {\color[HTML]{000000} 62.3}          & {\color[HTML]{000000} 73.2}          \\
\multirow{-7}{*}{{\color[HTML]{000000} \begin{tabular}[c]{@{}c@{}}RetinaNet\\ResNet101→\\ ResNeXt101\end{tabular}}}  & {\color[HTML]{000000} \textbf{DFMSD(ours)}}                                       & {\color[HTML]{000000} \textbf{\begin{tabular}[c]{@{}c@{}}42.0 (+4.6)\end{tabular}}} & {\color[HTML]{000000} \textbf{24.1}} & {\color[HTML]{000000} \textbf{46.3}} & {\color[HTML]{000000} \textbf{56.5}} & {\color[HTML]{000000} \textbf{\begin{tabular}[c]{@{}c@{}}58.2\end{tabular}}} & {\color[HTML]{000000} \textbf{38.9}} & {\color[HTML]{000000} \textbf{62.9}} & {\color[HTML]{000000} \textbf{75.2}} \\ \hline
{\color[HTML]{000000} }                                                                                           & {\color[HTML]{000000} \begin{tabular}[c]{@{}c@{}}RepPoints (baseline)\end{tabular}}   & {\color[HTML]{000000} 38.6}                                                           & {\color[HTML]{000000} 22.5}          & {\color[HTML]{000000} 42.2}          & {\color[HTML]{000000} 50.4}          & {\color[HTML]{000000} 55.1}                                                           & {\color[HTML]{000000} 34.9}          & {\color[HTML]{000000} 59.4}          & {\color[HTML]{000000} 70.3}          \\
{\color[HTML]{000000} }                                                                                           & {\color[HTML]{000000} FKD}                                                          & {\color[HTML]{000000} \begin{tabular}[c]{@{}c@{}}40.6 (+2.0)\end{tabular}}          & {\color[HTML]{000000} 23.4}          & {\color[HTML]{000000} 44.6}          & {\color[HTML]{000000} 53.0}          & {\color[HTML]{000000} \begin{tabular}[c]{@{}c@{}}56.9\end{tabular}}          & {\color[HTML]{000000} 37.3}          & {\color[HTML]{000000} 60.9}          & {\color[HTML]{000000} 71.4}          \\
{\color[HTML]{000000} }                                                                                           & {\color[HTML]{000000} FGD}                                                          & {\color[HTML]{000000} \begin{tabular}[c]{@{}c@{}}42.0 (+3.4)\end{tabular}}          & {\color[HTML]{000000} 24.0}          & {\color[HTML]{000000} 45.7}          & {\color[HTML]{000000} 55.6}          & {\color[HTML]{000000} \begin{tabular}[c]{@{}c@{}}58.2\end{tabular}}          & {\color[HTML]{000000} 37.8}          & {\color[HTML]{000000} 62.2}          & {\color[HTML]{000000} 73.3}          \\
{\color[HTML]{000000} }                                                                                           & {\color[HTML]{000000} MGD}                                                          & {\color[HTML]{000000} \begin{tabular}[c]{@{}c@{}}42.3 (+3.7)\end{tabular}}          & {\color[HTML]{000000} 24.4}          & {\color[HTML]{000000} 46.2}          & {\color[HTML]{000000} 55.9}          & {\color[HTML]{000000} \begin{tabular}[c]{@{}c@{}}58.4\end{tabular}}          & {\color[HTML]{000000} 40.4}          & {\color[HTML]{000000} 62.3}          & {\color[HTML]{000000} 73.9}          \\
{\color[HTML]{000000} }                                                                                           & {\color[HTML]{000000} AMD}                                                          & {\color[HTML]{000000} \begin{tabular}[c]{@{}c@{}}42.7 (+4.1)\end{tabular}}          & {\color[HTML]{000000} 24.8}          & {\color[HTML]{000000} 46.5}          & {\color[HTML]{000000} 56.3}          & {\color[HTML]{000000} \begin{tabular}[c]{@{}c@{}}58.8\end{tabular}}          & {\color[HTML]{000000} 40.6}          & {\color[HTML]{000000} 62.4}          & {\color[HTML]{000000} 74.1}          \\
{\color[HTML]{000000} }                                                                                           & {\color[HTML]{000000} DMKD}                                                         & {\color[HTML]{000000} \begin{tabular}[c]{@{}c@{}}42.9 (+4.3)\end{tabular}}          & {\color[HTML]{000000} 25.1}          & {\color[HTML]{000000} 46.9}          & {\color[HTML]{000000} 56.4}          & {\color[HTML]{000000} \begin{tabular}[c]{@{}c@{}}60.1\end{tabular}}          & {\color[HTML]{000000} 40.9}          & {\color[HTML]{000000} 62.9}          & {\color[HTML]{000000} 74.4}          \\
\multirow{-7}{*}{{\color[HTML]{000000} \begin{tabular}[c]{@{}c@{}}RepPoints \\ ResNet101→\\ ResNeXt101\end{tabular}}} & {\color[HTML]{000000} \textbf{DFMSD(ours)}}                                       & {\color[HTML]{000000} \textbf{\begin{tabular}[c]{@{}c@{}}43.2 (+4.6)\end{tabular}}} & {\color[HTML]{000000} \textbf{25.3}} & {\color[HTML]{000000} \textbf{47.4}} & {\color[HTML]{000000} \textbf{57.5}} & {\color[HTML]{000000} \textbf{\begin{tabular}[c]{@{}c@{}}60.2\end{tabular}}} & {\color[HTML]{000000} \textbf{41.0}} & {\color[HTML]{000000} \textbf{63.4}} & {\color[HTML]{000000} \textbf{74.8}} \\ \hline
{\color[HTML]{000000} }                                                                                     & {\color[HTML]{000000} \begin{tabular}[c]{@{}c@{}}GFL (baseline)\end{tabular}}  & {\color[HTML]{000000} 39.6}                                                           & {\color[HTML]{000000} 22.3}          & {\color[HTML]{000000} 43.6}          & {\color[HTML]{000000} 52.2}          & {\color[HTML]{000000} 58.2}                                                           & {\color[HTML]{000000} 35.8}          & {\color[HTML]{000000} 63.1}          & {\color[HTML]{000000} 75.2}          \\
{\color[HTML]{000000} }                                                                                     & {\color[HTML]{000000} MGD}                                                   & {\color[HTML]{000000} \begin{tabular}[c]{@{}c@{}}40.4 (+0.8)\end{tabular}}          & {\color[HTML]{000000} 22.8}          & {\color[HTML]{000000} 44.1}          & {\color[HTML]{000000} 53.7}          & {\color[HTML]{000000} \begin{tabular}[c]{@{}c@{}}59.0\end{tabular}}          & {\color[HTML]{000000} 37.0}          & {\color[HTML]{000000} 63.7}          & {\color[HTML]{000000} 76.3}          \\
{\color[HTML]{000000} }                                                                                     & {\color[HTML]{000000} AMD}                                                   & {\color[HTML]{000000} \begin{tabular}[c]{@{}c@{}}40.5 (+0.9)\end{tabular}}          & {\color[HTML]{000000} 23.3}          & {\color[HTML]{000000} 44.2}          & {\color[HTML]{000000} 53.5}          & {\color[HTML]{000000} \begin{tabular}[c]{@{}c@{}}59.0\end{tabular}}          & {\color[HTML]{000000} 37.3}          & {\color[HTML]{000000} 63.6}          & {\color[HTML]{000000} 75.8}          \\
{\color[HTML]{000000} }                                                                                     & {\color[HTML]{000000} DMKD}                                                  & {\color[HTML]{000000} \begin{tabular}[c]{@{}c@{}}40.8 (+1.2)\end{tabular}}          & {\color[HTML]{000000} 23.3}          & {\color[HTML]{000000} 44.6}          & {\color[HTML]{000000} 53.7}          & {\color[HTML]{000000} \begin{tabular}[c]{@{}c@{}}59.3\end{tabular}}          & {\color[HTML]{000000} 37.9}          & {\color[HTML]{000000} 64.1}          & {\color[HTML]{000000} 76.4}          \\
\multirow{-5}{*}{{\color[HTML]{000000} \begin{tabular}[c]{@{}c@{}}GFL\\ ResNet101→\\ ResNeXt101\end{tabular}}}  & {\color[HTML]{000000} \textbf{DFMSD(ours)}}                                & {\color[HTML]{000000} \textbf{\begin{tabular}[c]{@{}c@{}}42.2 (+2.6)\end{tabular}}} & {\color[HTML]{000000} \textbf{23.6}} & {\color[HTML]{000000} \textbf{46.2}} & {\color[HTML]{000000} \textbf{55.3}} & {\color[HTML]{000000} \textbf{\begin{tabular}[c]{@{}c@{}}60.2\end{tabular}}} & {\color[HTML]{000000} \textbf{39.4}} & {\color[HTML]{000000} \textbf{64.9}} & {\color[HTML]{000000} \textbf{76.5}} \\ \hline
{\color[HTML]{000000} }                                                                                     & {\color[HTML]{000000} \begin{tabular}[c]{@{}c@{}}FCOS (baseline)\end{tabular}} & {\color[HTML]{000000} 35.3}                                                           & {\color[HTML]{000000} 20.1}          & {\color[HTML]{000000} 38.3}          & {\color[HTML]{000000} 46.2}          & {\color[HTML]{000000} 53.0}                                                           & {\color[HTML]{000000} 32.3}          & {\color[HTML]{000000} 57.4}          & {\color[HTML]{000000} 69.4}          \\
{\color[HTML]{000000} }                                                                                     & {\color[HTML]{000000} MGD}                                                   & {\color[HTML]{000000} \begin{tabular}[c]{@{}c@{}}36.3 (+1.0)\end{tabular}}          & {\color[HTML]{000000} 20.3}          & {\color[HTML]{000000} 39.2}          & {\color[HTML]{000000} 48.0}          & {\color[HTML]{000000} \begin{tabular}[c]{@{}c@{}}53.7\end{tabular}}          & {\color[HTML]{000000} 32.8}          & {\color[HTML]{000000} 58.0}          & {\color[HTML]{000000} 70.3}          \\
{\color[HTML]{000000} }                                                                                     & {\color[HTML]{000000} AMD}                                                   & {\color[HTML]{000000} \begin{tabular}[c]{@{}c@{}}36.6 (+1.3)\end{tabular}}          & {\color[HTML]{000000} 20.5}          & {\color[HTML]{000000} 39.6}          & {\color[HTML]{000000} 48.5}          & {\color[HTML]{000000} \begin{tabular}[c]{@{}c@{}}54.0\end{tabular}}          & {\color[HTML]{000000} 33.1}          & {\color[HTML]{000000} 58.4}          & {\color[HTML]{000000} 70.9}          \\
{\color[HTML]{000000} }                                                                                     & {\color[HTML]{000000} DMKD}                                                  & {\color[HTML]{000000} \begin{tabular}[c]{@{}c@{}}36.9 (+1.6)\end{tabular}}          & {\color[HTML]{000000} 20.8}          & {\color[HTML]{000000} 40.0}          & {\color[HTML]{000000} 48.8}          & {\color[HTML]{000000} \begin{tabular}[c]{@{}c@{}}54.2\end{tabular}}          & {\color[HTML]{000000} 33.3}          & {\color[HTML]{000000} 58.8}          & {\color[HTML]{000000} 71.1}          \\
\multirow{-5}{*}{{\color[HTML]{000000} \begin{tabular}[c]{@{}c@{}}FCOS\\ ResNet101→\\ ResNeXt101\end{tabular}}} & {\color[HTML]{000000} \textbf{DFMSD(ours)}}                                & {\color[HTML]{000000} \textbf{\begin{tabular}[c]{@{}c@{}}37.2 (+1.9)\end{tabular}}} & {\color[HTML]{000000} \textbf{20.6}} & {\color[HTML]{000000} \textbf{40.2}} & {\color[HTML]{000000} \textbf{49.0}} & {\color[HTML]{000000} \textbf{\begin{tabular}[c]{@{}c@{}}54.8\end{tabular}}} & {\color[HTML]{000000} \textbf{34.7}} & {\color[HTML]{000000} \textbf{58.7}} & {\color[HTML]{000000} \textbf{71.2}} \\ \hline
\end{tabular}
\end{table*}

\subsection{Ablation Studies}\label{subsec46}

In this section, extensive ablation experiments are conducted to gain a deeper insight into different module and configurations on the performance of our proposed distillation framework. Similar to the settings in the above-mentioned experiments, various ViT and CNN detectors are involved in our ablation studies.

\begin{table*}[h] 
\centering
\caption{Ablating studies of SAL module within our proposed framework. The models in brackets indicate the backbone networks of the teacher detectors. The student framework is RetinaNet with ResNet50 backbone. The best result is achieved when the stage-wise Cascade Mask R-CNN teacher detectors successively utilize ResNet101 and ResNeXt101 as the backbones.}
\label{tab7}
\setlength{\tabcolsep}{3.5mm}
\begin{tabular}{c|c|c}
\hline
Teacher                                                                        & Student                                                                                 & mAP           \\ \hline
Cascade Mask R-CNN (ResNet50 $\rightarrow$ ResNet101 $\rightarrow$ ResNeXt101) & \multirow{6}{*}{\begin{tabular}[c]{@{}c@{}}RetinaNet-ResNet50 \\ (mAP: 37.4) \end{tabular}} & 40.1          \\
Cascade Mask R-CNN (ResNet50 $\rightarrow$ ResNet101)                          &                                                                                         & 39.8          \\
\textbf{Cascade Mask R-CNN (ResNet101 $\rightarrow$ ResNeXt101)}               &                                                                                         & \textbf{40.1} \\
Cascade Mask R-CNN (ResNet101) $\rightarrow$ FCOS (ResNeXt101)                 &                                                                                         & 37.6          \\
Cascade Mask R-CNN (ResNet101) $\rightarrow$ RetinaNet (ResNeXt101)            &                                                                                         & 38.6          \\
Cascade Mask R-CNN (ResNet101) $\rightarrow$ ST-T                              &                                                                                         & 40.0          \\ \hline
\end{tabular}
\end{table*}

\subsubsection{SAL module}
We have carried out different groups of experiments to explore the effect of distillation stages and different teacher detection frameworks on the model performance. More specifically, the teacher detectors include Cascade Mask R-CNN, FCOS, RetinaNet, and ST-T while RetinaNet with ResNet50 is used as the student model. As illustrated in Table~\ref{tab7}, the highest 40.1\% mAP accuracy is reported when the student successively learns from the Cascade mask R-CNN with ResNet101 and ResNext101 backbones. Interestingly, this result is even identical to the case when three teachers with successive ResNet50, ResNet101, and ResNeXt101 backbone networks are incorporated into our SAL module, which suggests that excessive distillation stages may not benefit improving the student performance due to the limitation of the representation power of similar teacher models. In addition, it is shown that deteriorating performance is reported when the teacher and the student detectors have diverse network architectures. For example, when the Cascade Mask R-CNN remains the ``weaker'' teacher framework and the ``stronger'' counterpart is used as an even more advanced ST-T framework, a slightly lower 40.0\% mAP score is achieved, which lags behind the case when both teacher models simultaneously utilize the Cascade Mask R-CNN framework. This implies that the gap among multiple teacher models may be detrimental to the distillation performance. 

\begin{table}[h] 
\centering
\caption{Ablation studies of our masking enhancement module during different distillation stages within our framework. The best result is achieved when imposing masking enhancement on the ``stronger'' teacher for the second distillation phase.}
\label{tab8}
\setlength{\tabcolsep}{3.5mm}
\begin{tabular}{cc|c}
\hline
{\color[HTML]{000000} }                                                                                            & {\color[HTML]{000000} }                                                                                             & {\color[HTML]{000000} \begin{tabular}[c]{@{}c@{}}Student:\\ RetinaNet-ResNet50\end{tabular}} \\ \cline{3-3} 
\multirow{-3}{*}{{\color[HTML]{000000} \begin{tabular}[c]{@{}c@{}}Cascade\\ Mask R-CNN\\ (ResNet101)\end{tabular}}} & \multirow{-3}{*}{{\color[HTML]{000000} \begin{tabular}[c]{@{}c@{}}Cascade\\ Mask R-CNN\\ (ResNeXt101)\end{tabular}}} & mAP                                                                                   \\ \hline
\checkmark                                                                                                                  & \checkmark                                                                                                                   & 40.0                                                                                  \\
                                                                                                            \checkmark       &                                                                                                                    & 39.6                                                                                  \\
                                                                                                                 & \textbf{}      \checkmark                                                                                                      & \textbf{40.1}                                                                         \\ \hline
\end{tabular}
\end{table}

\subsubsection{Masking enhancement module}
To explore the effect of the masking enhancement (ME) module on different distillation stages within our SAL module, we conduct a series of experiments in which the module is introduced into the first stage, the second stage, and both stages simultaneously. Specifically, the teacher Cascade Mask R-CNN detector successively leverages ResNet101 and ResNeXt101 for backbone networks and the RetinaNet-ResNet50 is used as the student model. As revealed in Table~\ref{tab8}, integrating the masking enhancement module in both stages can not bring further performance improvement, since extra enhancement may generate repeatedly identified object-aware regions, and thus produce biased detection results. In contrast, our method achieves slightly superior performance of 40.1\% mAP by only introducing masking enhancement into the second distillation stage. This suggests that ``stronger'' teacher with more powerful representation capability can benefit from the masking enhancement for better identifying the enhanced object-aware regions.

\begin{table}[h] 
\centering
\caption{Ablation studies of semantic alignment modules within our model. P1, P2 and P3 denote the respective FPN layers of the teacher and the student models.}
\label{tab9}
\setlength{\tabcolsep}{3.5mm}
\begin{tabular}{cccc}
\hline
\multicolumn{4}{c}{{\color[HTML]{000000} \begin{tabular}[c]{@{}c@{}}Teacher: Cascade Mask R-CNN (ResNet101$\rightarrow$ResNeXt101)\\ Student: RetinaNet-ResNet50\end{tabular}}} \\ \hline
P1                           & P2                           & \multicolumn{1}{c|}{P3}                           & mAP                            \\ \hline
                             &                              & \multicolumn{1}{c|}{\checkmark}                            & 39.8                           \\
                             & \checkmark                            & \multicolumn{1}{c|}{\checkmark}                            & 40.0                           \\
\checkmark                            & \checkmark                            & \multicolumn{1}{c|}{\checkmark}                            & \textbf{40.1}  \\ \hline                        
\end{tabular}
\end{table}

\subsubsection{Semantic Feature Alignment module}
To investigate the effect of Semantic Feature Alignment (SFA) module on the model performance, we perform semantic alignment at different feature layers between the teacher and the student backbone networks within our DFMSD model using different configurations. Consistent with the aforementioned setup, the Cascade Mask R-CNN with ResNet101 and ResNeXt101 backbones are used as the dual teachers and the student detector is RetinaNet-ResNet50. As shown in Table~\ref{tab9}, performing semantic alignment at each FPN layer from P1 to P3 between the teacher and the student helps to generate consistent feature distribution and thus achieves the best result. This also indicates that the teacher-student gap is manifested in the variance in feature distribution at each feature layer. 


\begin{table}[h] 
\centering
\caption{Comprehensive ablation studies for heterogeneous teacher and student detectors.}
\label{tab10}
\setlength{\tabcolsep}{3.5mm}
\begin{tabular}{cccc}
\hline
\multicolumn{4}{c}{\begin{tabular}[c]{@{}c@{}}Teacher: ST-T$\rightarrow$ST-S\\ Student: RetinaNet-ResNet50\end{tabular}} \\ \hline
SAL               & ME               & \multicolumn{1}{c|}{SFA}                    & mAP                        \\ \hline
\checkmark                 &                   & \multicolumn{1}{c|}{}                        & 40.8                       \\
                  & \checkmark                 & \multicolumn{1}{c|}{}                        & 40.5                       \\
                  &                   & \multicolumn{1}{c|}{\checkmark}                       & 40.2                       \\
\checkmark                 & \checkmark                 & \multicolumn{1}{c|}{\textbf{}}               & 41.1                       \\
\checkmark                 &                   & \multicolumn{1}{c|}{\checkmark}                       & 40.9                       \\
                  & \checkmark                 & \multicolumn{1}{c|}{\checkmark}                       & 40.7                       \\
\checkmark                 & \checkmark                 & \multicolumn{1}{c|}{\checkmark}                       & \textbf{41.2}              \\ \hline
\end{tabular}
\end{table}

\subsubsection{Ablating each module within our DFMSD framework}
In this section, we have comprehensively explored the three modules mentioned above by ablating each one in our experiments. The ablation studies fall into two groups according to the distillation setting, namely heterogeneous and homogeneous distillation. For heterogeneous distillation, the teachers are Transformer-based ST-T and ST-S models with the student detector used as RetinaNet-ResNet50. As demonstrated in Table~\ref{tab10}, suboptimal result is reported when any one module operates independently. In particular, when a single SAL produces promising 40.8\% mAP, combining it with ME and SFA modules further improves from 40.8\% to 41.2\%, substantially suggesting the benefits of integrating the complementary modules into the dual masking feature distillation framework. Similar results are also obtained in the ablation studies for homogeneous distillation where RetinaNet-ResNet101 and RetinaNet-ResNeXt101 are teacher models, while RetinaNet-ResNet50 is the student counterpart, demonstrating that the highest result of 42.0\% mAP is obtained when all three modules are integrated as shown in Table~\ref{tab11}.

\begin{table}[h] 
\centering
\caption{Comprehensive ablation studies for homogeneous teacher and student detectors.}
\label{tab11}
\setlength{\tabcolsep}{3.5mm}
\begin{tabular}{cccc}
\hline
\multicolumn{4}{c}{{\color[HTML]{000000} \begin{tabular}[c]{@{}c@{}}Teacher: RetinaNet (ResNet101$\rightarrow$ResNeXt101)\\ Student: RetinaNet-ResNet50\end{tabular}}} \\ \hline
SAL                     & ME                     & \multicolumn{1}{c|}{SFA}                         & mAP                              \\ \hline
\checkmark                       &                         & \multicolumn{1}{c|}{}                             & 41.7                             \\
                        & \checkmark                       & \multicolumn{1}{c|}{}                             & 41.4                             \\
                        &                         & \multicolumn{1}{c|}{\checkmark}                            & 41.1                             \\
\checkmark                       & \checkmark                       & \multicolumn{1}{c|}{\textbf{}}                    & 41.9                             \\
\checkmark                       &                         & \multicolumn{1}{c|}{\checkmark}                            & 41.8                             \\
                        & \checkmark                       & \multicolumn{1}{c|}{\checkmark}                            & 41.6                             \\
\checkmark                       & \checkmark                       & \multicolumn{1}{c|}{\checkmark}                            & \textbf{42.0}                    \\ \hline
\end{tabular}
\end{table}


\subsection{Parameter Analysis}\label{subsec47}
\begin{figure}[h]
\centering
\includegraphics[width=80mm]{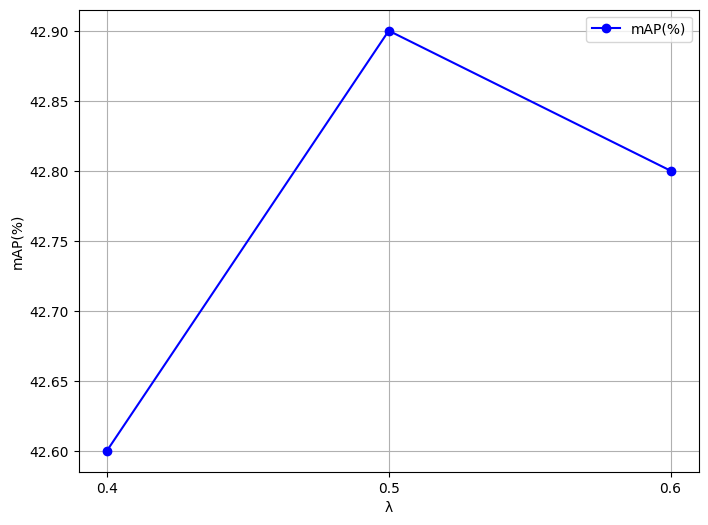}
\caption{Parameter analysis of $\lambda$ within our model. In this experiment, Faster R-CNN (ResNet50) is used as the student model and ST-T and ST-S are successively employed as ``weaker'' and ``stronger'' teacher models.}
\label{fig9}
\end{figure}

\begin{figure}[h]
\centering
\includegraphics[width=80mm]{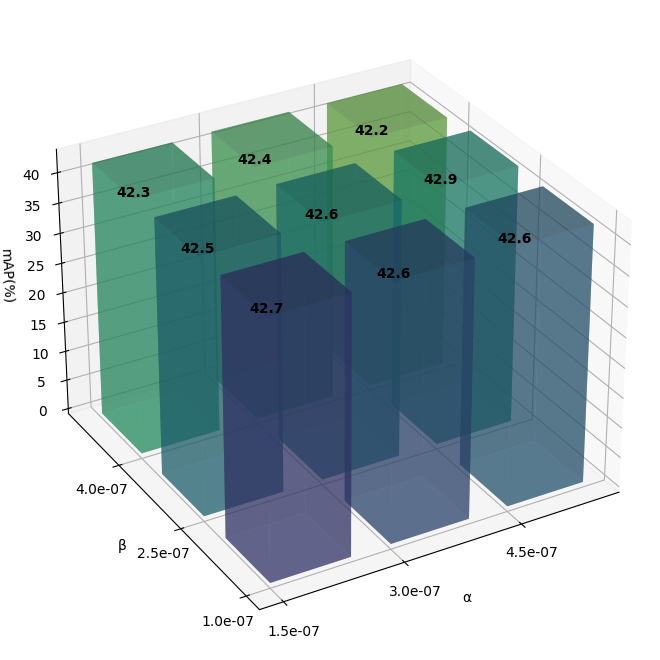}
\caption{Parameter analysis of $\alpha$ and $\beta$ within our model. The experimental setup follows that in Fig.~\ref{fig9}.}
\label{fig10}
\end{figure}

In this section, we discuss the setup of the hyperparameters involved in our DFMSD model. Firstly, various experimental evaluations are carried out using different threshold values $\lambda$, which indicates the scale distribution characteristics of the object-aware regions in Eq.~(\ref{eq4}). As shown in Fig.~\ref{fig9}, the best result is achieved when $\lambda=0.5$. This is reasonable since it is very likely that an image contains smaller objects when object-aware region areas account for less than half of the image size. In contrast, an image may constitute larger objects if $\lambda>0.5$.
In addition, we explore the impact of the hyperparameters $\alpha$ and $\beta$ in Eqs.~(\ref{eq6}) and (\ref{eq7}) on the model performance. As shown in Fig.~\ref{fig10}, it is shown that the highest 42.9\% mAP accuracy is achieved when $\alpha$ and $\beta$ are respectively set to 5.0×$10^{-7}$ and 2.5×$10^{-7}$, suggesting that different terms are balanced for desirable tradeoff.


\section{Conclusion}\label{sec5}

In this study, we have proposed a dual feature masking stage-wise distillation paradigm termed DFMSD to address heterogeneous distillation. More specifically, we propose integrating stage-wise learning into the dual feature masking framework such that the student can be progressively adapted to different teachers in various distillation stages. Meanwhile, masking enhancement is also introduced into the stage-wise learning such that the object-aware masking regions are enhanced for improved masking feature reconstruction. In addition, semantic alignment is also performed at different FPN layers between the teacher and the student network to generate consistent feature distributions. With all the above-mentioned modules incorporated, the gap between the teacher and the student models can be bridged for boosted distillation performance. Extensive experiments in the COCO dataset for object detection with different setups demonstrate the promise of our proposed method and the superiority to the SOTA, particularly in the heterogeneous distillation scenario.

\section*{Acknowledgement}
The authors greatly appreciate the valuable and constructive comments of the editors and all the anonymous reviewers. This work was supported by the National Natural Science Foundation of China under Grant 62173186, 62076134, 62303230 and Jiangsu provincial colleges of Natural Science General Program under Grant 22KJB510004. 

\section*{Declarations}

The authors declare that they have no known competing financial interests or personal relationships that could have appeared to
influence the work reported in this paper.

\textbf{Conflict of Interests. }

The authors declare that they have no conflict of interest.

\textbf{Data availability statement. }

Data sharing not applicable to this article as no datasets were generated or analyzed during the current study.



\bibliographystyle{elsarticle-num} 
\bibliography{ref}

\begin{thebibliography}{10}
\expandafter\ifx\csname url\endcsname\relax
  \def\url#1{\texttt{#1}}\fi
\expandafter\ifx\csname urlprefix\endcsname\relax\def\urlprefix{URL }\fi
\expandafter\ifx\csname href\endcsname\relax
  \def\href#1#2{#2} \def\path#1{#1}\fi

\bibitem{r2}
G.~Hinton, O.~Vinyals, J.~Dean, Distilling the knowledge in a neural network, arXiv preprint arXiv:1503.02531 (2015).

\bibitem{r5}
R.~Adriana, B.~Nicolas, K.~S. Ebrahimi, C.~Antoine, G.~Carlo, B.~Yoshua, Fitnets: Hints for thin deep nets, Proc. ICLR 2~(3) (2015) 1.

\bibitem{r11}
G.~Chen, W.~Choi, X.~Yu, T.~Han, M.~Chandraker, Learning efficient object detection models with knowledge distillation, Advances in neural information processing systems 30 (2017).

\bibitem{r64}
Y.~Liu, K.~Chen, C.~Liu, Z.~Qin, Z.~Luo, J.~Wang, Structured knowledge distillation for semantic segmentation, in: Proceedings of the IEEE/CVF conference on computer vision and pattern recognition, 2019, pp. 2604--2613.

\bibitem{r74}
B.~Heo, M.~Lee, S.~Yun, J.~Y. Choi, Knowledge transfer via distillation of activation boundaries formed by hidden neurons, in: Proceedings of the AAAI conference on artificial intelligence, Vol.~33, 2019, pp. 3779--3787.

\bibitem{r75}
J.~Gou, B.~Yu, S.~J. Maybank, D.~Tao, Knowledge distillation: A survey, International Journal of Computer Vision 129~(6) (2021) 1789--1819.

\bibitem{r17}
H.~Zhou, L.~Song, J.~Chen, Y.~Zhou, G.~Wang, J.~Yuan, Q.~Zhang, Rethinking soft labels for knowledge distillation: A bias-variance tradeoff perspective, arXiv preprint arXiv:2102.00650 (2021).

\bibitem{r4}
W.~Park, D.~Kim, Y.~Lu, M.~Cho, Relational knowledge distillation, in: Proceedings of the IEEE/CVF conference on computer vision and pattern recognition, 2019, pp. 3967--3976.

\bibitem{r76}
S.~Zagoruyko, N.~Komodakis, Paying more attention to attention: Improving the performance of convolutional neural networks via attention transfer, arXiv preprint arXiv:1612.03928 (2016).

\bibitem{r77}
F.~Tung, G.~Mori, Similarity-preserving knowledge distillation, in: Proceedings of the IEEE/CVF international conference on computer vision, 2019, pp. 1365--1374.

\bibitem{r78}
B.~Heo, J.~Kim, S.~Yun, H.~Park, N.~Kwak, J.~Y. Choi, A comprehensive overhaul of feature distillation, in: Proceedings of the IEEE/CVF international conference on computer vision, 2019, pp. 1921--1930.

\bibitem{r19}
G.~Yang, Y.~Tang, Z.~Wu, J.~Li, J.~Xu, X.~Wan, Dmkd: Improving feature-based knowledge distillation for object detection via dual masking augmentation, in: ICASSP 2024-2024 IEEE International Conference on Acoustics, Speech and Signal Processing (ICASSP), IEEE, 2024, pp. 3330--3334.

\bibitem{r101}
S.~Guo, X.~Li, P.~Zhu, Z.~Mu, Ads-detector: An attention-based dual stream adversarial example detection method, Knowledge-Based Systems 265 (2023) 110388.

\bibitem{r79}
B.~Heo, J.~Kim, S.~Yun, H.~Park, N.~Kwak, J.~Y. Choi, A comprehensive overhaul of feature distillation, in: Proceedings of the IEEE/CVF international conference on computer vision, 2019, pp. 1921--1930.

\bibitem{r8}
T.-Y. Lin, P.~Goyal, R.~Girshick, K.~He, P.~Doll{\'a}r, Focal loss for dense object detection, in: Proceedings of the IEEE international conference on computer vision, 2017, pp. 2980--2988.

\bibitem{r29}
K.~He, G.~Gkioxari, P.~Doll{\'a}r, R.~Girshick, Mask r-cnn, in: Proceedings of the IEEE international conference on computer vision, 2017, pp. 2961--2969.

\bibitem{r80}
S.~I. Mirzadeh, M.~Farajtabar, A.~Li, N.~Levine, A.~Matsukawa, H.~Ghasemzadeh, Improved knowledge distillation via teacher assistant, in: Proceedings of the AAAI conference on artificial intelligence, Vol.~34, 2020, pp. 5191--5198.

\bibitem{r10}
S.~Ren, K.~He, R.~Girshick, J.~Sun, Faster r-cnn: Towards real-time object detection with region proposal networks, IEEE transactions on pattern analysis and machine intelligence 39~(6) (2016) 1137--1149.

\bibitem{r20}
Z.~Tian, C.~Shen, H.~Chen, T.~He, Fcos: Fully convolutional one-stage object detection, in: IEEE/CVF International Conference on Computer Vision (ICCV), 2019, pp. 9626--9635.

\bibitem{r81}
K.~He, X.~Zhang, S.~Ren, J.~Sun, Deep residual learning for image recognition, in: Proceedings of the IEEE conference on computer vision and pattern recognition, 2016, pp. 770--778.

\bibitem{r102}
L.~Li, J.~Xu, Multi-label category enhancement fusion distillation based on variational estimation, Knowledge-Based Systems (2024) 112092.

\bibitem{r97}
W.~Zhou, Z.~Huang, C.~Wang, Y.~Chen, A multi-graph neural group recommendation model with meta-learning and multi-teacher distillation, Knowledge-Based Systems 276 (2023) 110731.

\bibitem{r98}
S.~Ahmad, Z.~Ullah, J.~Gwak, Multi-teacher cross-modal distillation with cooperative deep supervision fusion learning for unimodal segmentation, Knowledge-Based Systems 297 (2024) 111854.

\bibitem{r82}
S.~Gupta, J.~Hoffman, J.~Malik, Cross modal distillation for supervision transfer, in: Proceedings of the IEEE conference on computer vision and pattern recognition, 2016, pp. 2827--2836.

\bibitem{r83}
B.~Heo, J.~Kim, S.~Yun, H.~Park, N.~Kwak, J.~Y. Choi, A comprehensive overhaul of feature distillation, in: Proceedings of the IEEE/CVF international conference on computer vision, 2019, pp. 1921--1930.

\bibitem{r91}
Y.~Deng, W.~Bai, Y.~Jiang, Y.~Tang, Subgraph-based feature fusion models for semantic similarity computation in heterogeneous knowledge graphs, Knowledge-Based Systems 257 (2022) 109906.

\bibitem{r7}
Y.~Zhang, T.~Xiang, T.~M. Hospedales, H.~Lu, Deep mutual learning, in: Proceedings of the IEEE conference on computer vision and pattern recognition, 2018, pp. 4320--4328.

\bibitem{r84}
N.~Passalis, A.~Tefas, Learning deep representations with probabilistic knowledge transfer, in: Proceedings of the European Conference on Computer Vision (ECCV), 2018, pp. 268--284.

\bibitem{r85}
H.~Li, P.~Xiong, H.~Fan, J.~Sun, Dfanet: Deep feature aggregation for real-time semantic segmentation, in: Proceedings of the IEEE/CVF conference on computer vision and pattern recognition, 2019, pp. 9522--9531.

\bibitem{r25}
C.-Y. Fu, W.~Liu, A.~Ranga, A.~Tyagi, A.~C. Berg, Dssd: Deconvolutional single shot detector, arXiv preprint arXiv:1701.06659 (2017).

\bibitem{r26}
N.~Carion, F.~Massa, G.~Synnaeve, N.~Usunier, A.~Kirillov, S.~Zagoruyko, End-to-end object detection with transformers, in: European conference on computer vision, Springer, 2020, pp. 213--229.

\bibitem{r27}
S.~Gidaris, N.~Komodakis, Object detection via a multi-region and semantic segmentation-aware cnn model, in: Proceedings of the IEEE international conference on computer vision, 2015, pp. 1134--1142.

\bibitem{r28}
T.~Kong, A.~Yao, Y.~Chen, F.~Sun, Hypernet: Towards accurate region proposal generation and joint object detection, in: Proceedings of the IEEE conference on computer vision and pattern recognition, 2016, pp. 845--853.

\bibitem{r30}
J.~Redmon, A.~Farhadi, Yolov3: An incremental improvement, arXiv preprint arXiv:1804.02767 (2018).

\bibitem{r31}
Z.~Ge, S.~Liu, F.~Wang, Z.~Li, J.~Sun, Yolox: Exceeding yolo series in 2021, arXiv preprint arXiv:2107.08430 (2021).

\bibitem{r65}
H.~Law, J.~Deng, Cornernet: Detecting objects as paired keypoints, in: Proceedings of the European conference on computer vision (ECCV), 2018, pp. 734--750.

\bibitem{r66}
X.~Zhou, D.~Wang, P.~Kr{\"a}henb{\"u}hl, Objects as points, arXiv preprint arXiv:1904.07850 (2019).

\bibitem{r67}
K.~Duan, S.~Bai, L.~Xie, H.~Qi, Q.~Huang, Q.~Tian, Centernet: Keypoint triplets for object detection, in: Proceedings of the IEEE/CVF international conference on computer vision, 2019, pp. 6569--6578.

\bibitem{r24}
Z.~Liu, Y.~Lin, Y.~Cao, H.~Hu, Y.~Wei, Z.~Zhang, S.~Lin, B.~Guo, Swin transformer: Hierarchical vision transformer using shifted windows, in: Proceedings of the IEEE/CVF international conference on computer vision, 2021, pp. 10012--10022.

\bibitem{r33}
H.~Zhang, F.~Li, S.~Liu, L.~Zhang, H.~Su, J.~Zhu, L.~M. Ni, H.-Y. Shum, Dino: Detr with improved denoising anchor boxes for end-to-end object detection, arXiv preprint arXiv:2203.03605 (2022).

\bibitem{r34}
X.~Zhu, W.~Su, L.~Lu, B.~Li, X.~Wang, J.~Dai, Deformable detr: Deformable transformers for end-to-end object detection, arXiv preprint arXiv:2010.04159 (2020).

\bibitem{r87}
A.~G. Howard, M.~Zhu, B.~Chen, D.~Kalenichenko, W.~Wang, T.~Weyand, M.~Andreetto, H.~Adam, Mobilenets: Efficient convolutional neural networks for mobile vision applications, arXiv preprint arXiv:1704.04861 (2017).

\bibitem{r88}
M.~Sandler, A.~Howard, M.~Zhu, A.~Zhmoginov, L.-C. Chen, Mobilenetv2: Inverted residuals and linear bottlenecks, in: Proceedings of the IEEE conference on computer vision and pattern recognition, 2018, pp. 4510--4520.

\bibitem{r40}
Q.~Li, S.~Jin, J.~Yan, Mimicking very efficient network for object detection, in: Proceedings of the ieee conference on computer vision and pattern recognition, 2017, pp. 6356--6364.

\bibitem{r12}
X.~Dai, Z.~Jiang, Z.~Wu, Y.~Bao, Z.~Wang, S.~Liu, E.~Zhou, General instance distillation for object detection, in: Proceedings of the IEEE/CVF conference on computer vision and pattern recognition, 2021, pp. 7842--7851.

\bibitem{r15}
Z.~Yang, Z.~Li, X.~Jiang, Y.~Gong, Z.~Yuan, D.~Zhao, C.~Yuan, Focal and global knowledge distillation for detectors, in: Proceedings of the IEEE/CVF Conference on Computer Vision and Pattern Recognition, 2022, pp. 4643--4652.

\bibitem{r42}
Z.~Yang, Z.~Li, M.~Shao, D.~Shi, Z.~Yuan, C.~Yuan, Masked generative distillation, in: European Conference on Computer Vision, Springer, 2022, pp. 53--69.

\bibitem{r43}
G.~Yang, Y.~Tang, J.~Li, J.~Xu, X.~Wan, Amd: Adaptive masked distillation for object detection, in: 2023 International Joint Conference on Neural Networks (IJCNN), IEEE, 2023, pp. 1--8.

\bibitem{r35}
X.~Lu, Q.~Li, B.~Li, J.~Yan, Mimicdet: Bridging the gap between one-stage and two-stage object detection, in: Computer Vision--ECCV 2020: 16th European Conference, Glasgow, UK, August 23--28, 2020, Proceedings, Part XIV 16, Springer, 2020, pp. 541--557.

\bibitem{r36}
L.~Yao, R.~Pi, H.~Xu, W.~Zhang, Z.~Li, T.~Zhang, G-detkd: Towards general distillation framework for object detectors via contrastive and semantic-guided feature imitation, in: Proceedings of the IEEE/CVF international conference on computer vision, 2021, pp. 3591--3600.

\bibitem{r1}
L.~Wang, X.~Li, Y.~Liao, Z.~Jiang, J.~Wu, F.~Wang, C.~Qian, S.~Liu, Head: Hetero-assists distillation for heterogeneous object detectors, in: European Conference on Computer Vision, Springer, 2022, pp. 314--331.

\bibitem{r37}
W.~Cao, Y.~Zhang, J.~Gao, A.~Cheng, K.~Cheng, J.~Cheng, Pkd: General distillation framework for object detectors via pearson correlation coefficient, Advances in Neural Information Processing Systems 35 (2022) 15394--15406.

\bibitem{r38}
J.~Wang, Y.~Chen, Z.~Zheng, X.~Li, M.-M. Cheng, Q.~Hou, Crosskd: Cross-head knowledge distillation for dense object detection, arXiv preprint arXiv:2306.11369 (2023).

\bibitem{r99}
X.~Guo, W.~Zhou, T.~Liu, Contrastive learning-based knowledge distillation for rgb-thermal urban scene semantic segmentation, Knowledge-Based Systems 292 (2024) 111588.

\bibitem{r100}
H.~Cheng, X.~Han, P.~Shi, J.~Zhu, Z.~Li, Multi-trusted cross-modal information bottleneck for 3d self-supervised representation learning, Knowledge-Based Systems 283 (2024) 111217.

\bibitem{r45}
H.~Wang, X.~Wu, Z.~Huang, E.~P. Xing, High-frequency component helps explain the generalization of convolutional neural networks, in: Proceedings of the IEEE/CVF conference on computer vision and pattern recognition, 2020, pp. 8684--8694.

\bibitem{r93}
Z.~Dong, Q.~Hu, Z.~Zhang, J.~Zhao, On the effectiveness of graph data augmentation for source code learning, Knowledge-Based Systems 285 (2024) 111328.

\bibitem{r94}
X.~Li, Y.~Wu, C.~Tang, Y.~Fu, L.~Zhang, Improving generalization of convolutional neural network through contrastive augmentation, Knowledge-Based Systems 272 (2023) 110543.

\bibitem{r48}
K.~Maharana, S.~Mondal, B.~Nemade, A review: Data pre-processing and data augmentation techniques, Global Transitions Proceedings 3~(1) (2022) 91--99.

\bibitem{r49}
R.~Takahashi, T.~Matsubara, K.~Uehara, Data augmentation using random image cropping and patching for deep cnns, IEEE Transactions on Circuits and Systems for Video Technology 30~(9) (2019) 2917--2931.

\bibitem{r50}
E.~Kim, J.~Kim, H.~Lee, S.~Kim, Adaptive data augmentation to achieve noise robustness and overcome data deficiency for deep learning, Applied Sciences 11~(12) (2021) 5586.

\bibitem{r92}
M.~Wang, S.~Wang, Y.~Wang, W.~Wang, T.~Liang, J.~Chen, Z.~Luo, Boosting unsupervised domain adaptation: A fourier approach, Knowledge-Based Systems 264 (2023) 110325.

\bibitem{r58}
A.~Liu, X.~Liu, H.~Yu, C.~Zhang, Q.~Liu, D.~Tao, Training robust deep neural networks via adversarial noise propagation, IEEE Transactions on Image Processing 30 (2021) 5769--5781.

\bibitem{r89}
J.~Liang, S.~Liang, A.~Liu, K.~Ma, J.~Li, X.~Cao, Exploring inconsistent knowledge distillation for object detection with data augmentation, in: Proceedings of the 31st ACM International Conference on Multimedia, 2023, pp. 768--778.

\bibitem{r95}
C.~Xie, Z.~Zhang, Y.~Zhou, S.~Bai, J.~Wang, Z.~Ren, A.~L. Yuille, Improving transferability of adversarial examples with input diversity, in: Proceedings of the IEEE/CVF conference on computer vision and pattern recognition, 2019, pp. 2730--2739.

\bibitem{r96}
Y.~Wang, W.~Hong, X.~Zhang, Q.~Zhang, C.~Gu, Boosting transferability of adversarial samples via saliency distribution and frequency domain enhancement, Knowledge-Based Systems (2024) 112152.

\bibitem{r23}
T.-Y. Lin, M.~Maire, S.~Belongie, J.~Hays, P.~Perona, D.~Ramanan, P.~Doll{\'a}r, C.~L. Zitnick, Microsoft coco: Common objects in context, in: Computer Vision--ECCV 2014: 13th European Conference, Zurich, Switzerland, September 6-12, 2014, Proceedings, Part V 13, Springer, 2014, pp. 740--755.

\bibitem{r32}
R.~Girshick, Fast r-cnn, in: Proceedings of the IEEE international conference on computer vision, 2015, pp. 1440--1448.

\bibitem{r54}
X.~Li, W.~Wang, L.~Wu, S.~Chen, X.~Hu, J.~Li, J.~Tang, J.~Yang, Generalized focal loss: Learning qualified and distributed bounding boxes for dense object detection, in: Advances in Neural Information Processing Systems, 2020, pp. 21002--21012.

\bibitem{r9}
Z.~Yang, S.~Liu, H.~Hu, L.~Wang, S.~Lin, Reppoints: Point set representation for object detection, in: Proceedings of the IEEE/CVF international conference on computer vision, 2019, pp. 9657--9666.

\bibitem{r90}
M.~Everingham, L.~Van~Gool, C.~K. Williams, J.~Winn, A.~Zisserman, The pascal visual object classes (voc) challenge, International journal of computer vision 88 (2010) 303--338.

\bibitem{r53}
Z.~Shen, E.~Xing, A fast knowledge distillation framework for visual recognition, in: European Conference on Computer Vision, 2022, pp. 673--690.

\bibitem{r73}
X.~Li, W.~Wang, L.~Wu, S.~Chen, X.~Hu, J.~Li, J.~Tang, J.~Yang, Generalized focal loss: Learning qualified and distributed bounding boxes for dense object detection, Advances in Neural Information Processing Systems 33 (2020) 21002--21012.

\bibitem{r51}
K.~Chen, J.~Wang, J.~Pang, Y.~Cao, Y.~Xiong, X.~Li, S.~Sun, W.~Feng, Z.~Liu, J.~Xu, et~al., Mmdetection: Open mmlab detection toolbox and benchmark, arXiv preprint arXiv:1906.07155 (2019).

\end{thebibliography}





\end{document}